\documentclass{article}

\usepackage{iclr2027_conference,times}
\usepackage{amsmath,amsfonts,bm}

\def\eqref#1{equation~\ref{#1}}

\def\1{\bm{1}}

\DeclareMathAlphabet{\mathsfit}{\encodingdefault}{\sfdefault}{m}{sl}
\SetMathAlphabet{\mathsfit}{bold}{\encodingdefault}{\sfdefault}{bx}{n}

\usepackage[utf8]{inputenc}
\usepackage[T1]{fontenc}
\usepackage{xcolor}
\usepackage{hyperref}
\hypersetup{
    colorlinks=true, 
    linkcolor=red,
    citecolor=gray, 
    urlcolor=magenta,
    pdftitle={Zero-Shot Time-Series Question Answering via Decoupled Perception and Reasoning},
    pdfsubject={Time-series question answering},
    pdfkeywords={time series, question answering, zero-shot transfer, TSHarness}
}
\usepackage{url}
\usepackage{graphicx}
\usepackage{pgfplots}
\pgfplotsset{compat=1.18}
\usepackage{booktabs}
\usepackage{microtype}
\usepackage{amssymb}
\usepackage{mathtools}
\usepackage{multirow}
\usepackage{enumitem}
\setlist[itemize]{leftmargin=*,labelindent=0pt}

\newcommand{\modelname}{TSHarness}
\newcommand{\fullname}{\textsc{\modelname}}

\title{Zero-Shot Time-Series Question Answering via\\Decoupled Perception and Reasoning}

\author{Jing Xie\textsuperscript{1}\thanks{Equal contribution.}\quad Haochen Yuan\textsuperscript{1}\footnotemark[1]\quad Yunbo Wang\textsuperscript{1}\thanks{Corresponding author.}\\
\normalfont\small\textsuperscript{1}MoE Key Lab of Artificial Intelligence, Institute of AI,\\
\normalfont\small School of Computer Science, Shanghai Jiao Tong University, Shanghai, China}
\hypersetup{pdfauthor={Jing Xie, Haochen Yuan, Yunbo Wang}}

\usepackage{arxiv-preprint}

\begin{document}

\maketitle

\begin{abstract}

Time-series question answering (TSQA) requires grounding linguistic queries and diverse answer formats in complex numerical observations. However, existing methods heavily overfit to specific datasets and struggle to generalize when input series, question contexts, and answer requirements shift simultaneously. To address this challenge, we propose \textbf{TSHarness}, an agentic framework that establishes a decoupled workflow for cross-dataset zero-shot TSQA. At its core, TSHarness divides and conquers numerical perception and contextual reasoning via a structured \textit{Time-Series Perception State} (TPS). Guided by a reusable memory of analytical knowledge, a learned \textit{Tool Selector} adaptively invokes numerical tools to extract salient statistical and temporal features into the TPS. The answering agent then performs semantic reasoning over the TPS to generate target outputs, triggering iterative re-perception through the feedback loop when evidence is deemed insufficient. By separating numerical feature extraction from question-specific reasoning, TSHarness eliminates the need for target-side training or answer feedback, providing a generalizable, cost-efficient foundation for zero-shot TSQA.

\end{abstract}

\section{Introduction}
\label{sec:intro}

\begin{figure}[t]
\centering
\includegraphics[width=\textwidth]{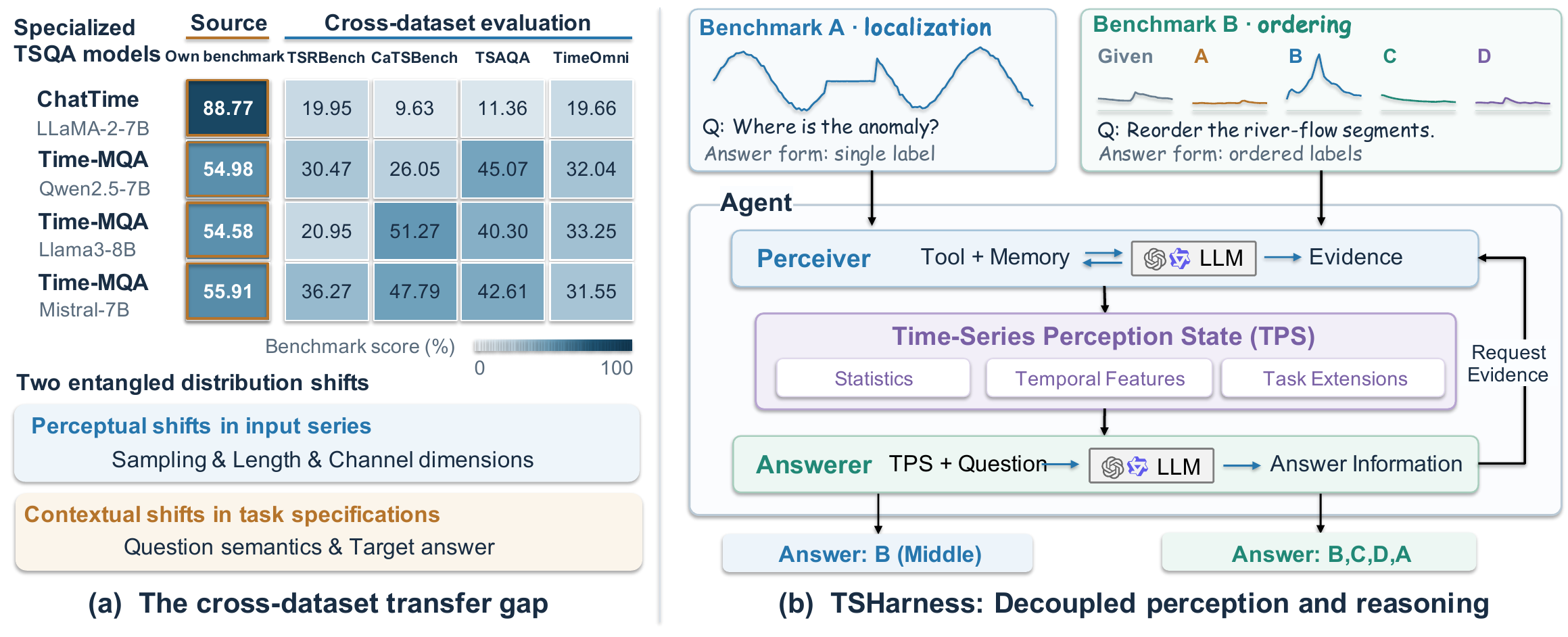}
\vspace{-20pt}
\caption{\textbf{Motivation and overview of \fullname{}.} (a) Specialized TSQA models exhibit uneven cross-dataset performance as time-series inputs, question context, and answer requirements change. (b) \fullname{} decouples perception from answering through a unified TPS representation. }
\vspace{-10pt}
\label{fig:intro}
\end{figure}

Time-series question answering (TSQA) turns numerical observations into answers about temporal patterns, events, predictions, and decisions~\citep{timemqa,tsrbench,timeomni}.
Current TSQA methods develop useful temporal capabilities through numerical-text alignment and task-specific training~\citep{chattime,timemqa}. Yet, we observe that performance gains of these models on one dataset do not reliably transfer to another (Figure~\ref{fig:intro}(a)), where perception, reasoning, and answer production must be recombined.

We attribute this transfer gap to prevailing paradigms that treat TSQA as a monolithic task, tightly entangling numerical perception with dataset-specific answer generation.
Consequently, such end-to-end models underperform in \textit{zero-shot} TSQA tasks under two entangled distribution shifts: \emph{perceptual shifts in input series} across sampling rates, sequence lengths, missingness, and channel dimensions; and (2) \emph{contextual shifts in task specifications} where question semantics and target answer requirements diverge together, spanning multiple-choice, true/false, and ordered labels under heterogeneous evaluation rubrics.
These concurrent shifts fundamentally impede knowledge reuse across domains: fine-tuned parameters of learning-based methods overfit to source dataset correlations (\emph{parameter reuse failure}), while stored analytical procedures and memories in agent-based methods fail to generalize across divergent benchmarks (\emph{analytical reuse failure}).

To overcome this dual transfer bottleneck, our core design principle is to decouple continuous numerical perception from context-specific reasoning.
We investigate this paradigm under the rigorous setting of cross-dataset zero-shot transfer, where a source-developed system directly answers target questions without any target-side fine-tuning or answer feedback.
Accordingly, we present \fullname{}, an agentic system centered on an intermediate representation termed the Time-Series Perception State (TPS), which explicitly isolates raw numerical observations from task-dependent answer synthesis.
Within this framework (Figure~\ref{fig:intro}(b)), a frozen backbone LLM instantiates two collaborative roles: a \textsc{Perceiver} that extracts quantitative temporal properties into the TPS, and an \textsc{Answerer} that conditions on the TPS to resolve the contextual question.

Specifically, the TPS establishes a unified analytical basis through standardized fields for quantitative temporal primitives, alongside flexible extensions for question-specific analyses.
During execution, guided by reusable procedural memory, the \textsc{Perceiver} dynamically invokes numerical tools to populate the TPS, translating raw series into domain-agnostic facts.
The \textsc{Answerer} then reasons over the accumulated TPS and question context to synthesize the final output.
Crucially, when existing evidence is deemed insufficient, the \textsc{Answerer} initiates an iterative feedback loop back to the \textsc{Perceiver}, requesting targeted, fine-grained measurements.
Furthermore, to mitigate redundant computational overhead, a lightweight reinforcement-learning (RL) \textsc{Tool Selector} adaptively prunes sub-optimal tool proposals by balancing answer correctness against execution latency on source tasks.
Through this decoupled architecture, source-developed analytical capabilities are dynamically recombined at runtime, enabling robust zero-shot generalization across unseen benchmarks while keeping the underlying backbone and analytical resources entirely fixed.

Our primary contributions are summarized as follows:
\begin{itemize}
\vspace{-5pt}
\item We systematically formulate the challenge of zero-shot TSQA, identifying that coupled perceptual and contextual shifts trigger distinct parameter and procedural reuse failures in existing methods.
\item We propose \fullname{}, which isolates numerical perception from contextual reasoning via the intermediate TPS. Guided by reusable memory, collaborative \textsc{Perceiver} and \textsc{Answerer} agents resolve queries through an active evidence loop.
\item We develop a lightweight RL-based approach to balance answer correctness against execution cost, substantially pruning redundant tool calls while maintaining competitive accuracy.
\item We benchmark zero-shot transfer of \fullname{} from a single source to five heterogeneous target datasets. It remarkably outperforms same-backbone baselines without any target-side adaptation.
\vspace{-5pt}
\end{itemize}

\section{Related Work}

Language-based forecasting~\citep{llmtime,timellm,unitime} and pretrained forecasters~\citep{chronos,moirai} support transfer of numerical prediction capabilities. TSQA broadens this scope to temporal reasoning~\citep{catsbench,tsaqa,tsrbench,tsexamagent} and contextual sensor interpretation~\citep{sensorqa}. Recent benchmarks reveal cross-domain weaknesses in time-series-adapted LLMs~\citep{mmtsbench} and distinguish scale selection, localization, and cross-interval integration~\citep{tsskill}. Learned approaches connect numerical observations to language through expanded vocabularies in ChatTime~\citep{chattime}, contextual QA training in Time-MQA~\citep{timemqa}, and synthetic alignment data in ChatTS~\citep{chatts}; TimeOmni-1 further develops reasoning through reward-guided training~\citep{timeomni}. \fullname{} makes statistical and temporal properties explicit through the TPS, separating their representation from question-specific interpretation. This provides a common analytical basis for applying a system constructed on one dataset to questions and answer requirements in others.

Chain-of-thought prompting structures intermediate reasoning~\citep{cot}; tool-augmented agents connect reasoning to execution~\citep{react,pal,toolformer} and reuse feedback, skills, and memory~\citep{reflexion,voyager,awm,amem}. Planning--execution separation~\citep{rewoo}, parallel scheduling~\citep{llmcompiler}, and modular architecture search~\citep{agentsquare} improve orchestration, while reward-based training develops LLM tool-use policies~\citep{toolrl,retool}. Time-series frameworks specialize these capabilities: TS-Agent combines numerical tools with evidence gathering and verification~\citep{tsagent}, AION organizes temporally grounded workflows and review~\citep{aion}, and KairosAgent learns reasoning--forecasting interactions from forecasting feedback~\citep{kairosagent}. Our closest harness baseline, TimeClaw, integrates temporal tools, capability evolution, and episodic retrieval~\citep{timeclaw}. \fullname{} focuses on the structure of evidence passed between perception and answering. TPS gives measured series properties a stable field structure and records task-specific calculations in separate extensions. This division lets the Answerer adapt its interpretation and evidence requests while retaining a common interface to numerical perception. A complementary RL selector learns to refine base-tool subsets with the backbone fixed, balancing answer correctness against tool executions.

\section{Method}
\label{sec:method}
\vspace{-5pt}

This section provides the technical details of \fullname{}. Sec.~\ref{sec:framework-overview} introduces our problem formulation and decoupled execution pipeline for cross-dataset zero-shot TSQA.
Sec.~\ref{sec:perception} and Sec.~\ref{sec:tps} detail our tool-augmented numerical perception and establish the Time-Series Perception State (TPS), which isolates objective temporal properties from task-specific reasoning.
Sec.~\ref{sec:answering} establishes an active harness loop for iterative evidence reasoning. Sec.~\ref{sec:selector} presents an RL-based \textsc{Tool Selector} that balances answer correctness against tool execution cost.

\subsection{Framework Overview}
\label{sec:framework-overview}

\begin{figure}[t]
\centering
\includegraphics[width=\textwidth]{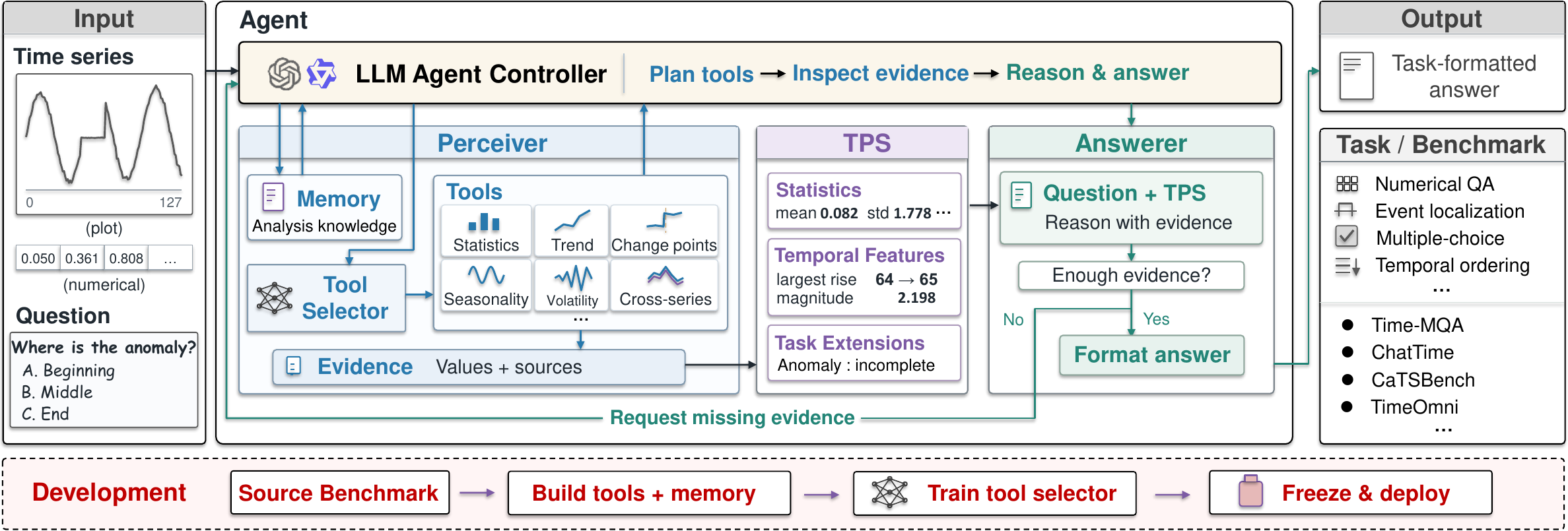}
\vspace{-18pt}
\caption{\textbf{Workflow of \fullname{}.} Guided by procedural memory, the \textsc{Perceiver} populates the TPS with grounded measurements, refined by the \textsc{Tool Selector}. The \textsc{Answerer} interprets the accumulated evidence alongside the question, actively initiating an iterative feedback loop to request further evidence. Source-developed resources are completely frozen for zero-shot transfer.}
\vspace{-5pt}
\label{fig:method}
\end{figure}

\fullname{} builds on a foundational insight: while questions across heterogeneous TSQA benchmarks differ drastically in textual framing and answer specifications, they fundamentally rely on a shared set of underlying statistical and temporal properties.
Formally, let a TSQA query instance be denoted as $u=(X, c, \kappa)$, where $X=(X_1,\dots,X_m)$ comprises the observed numerical series along with their metadata descriptions, $c$ denotes the question text and relevant background context, and $\kappa$ encapsulates the target answer specifications (\textit{e.g.}, categorical options, numerical tolerance, or formatting rubrics).
Let $Z_r$ denote the TPS representation after $r$ accepted evidence refinement updates. The overall execution workflow of \fullname{} is formulated as:
\begin{equation}
Z_0=\operatorname{Init}(u),\qquad
Z_{r+1}=\operatorname{Update}(Z_r,q_r;X),\qquad
\hat y=\operatorname{Answer}(Z_R,c,\kappa).
\label{eq:framework-flow}
\end{equation}
where $\operatorname{Init}(\cdot)$ constructs $Z_0$ from base measurements and question-specific analyses under the \textsc{Perceiver}'s validated plan, guided by procedural memory and optionally refined by the \textsc{Tool Selector}.
At iteration $r$, an evidence request $q_r$ dispatched by the \textsc{Answerer} invokes specific numerical operations, appending newly verified empirical findings to update the state from $Z_r$ to $Z_{r+1}$.
The loop terminates at step $R$ when the accumulated evidence is deemed sufficient or the execution budget is reached. $\operatorname{Answer}(\cdot)$ then synthesizes the final prediction $\hat y$ conforming to $\kappa$.
A single frozen backbone LLM supports planning and answering through role-specific calls, coordinated by the harness loop in Figure~\ref{fig:method}.

We construct the system exclusively on a single source dataset, using execution trajectories and ground-truth answer feedback to curate the tool library, analytical memory bank, and policy parameters of the \textsc{Tool Selector}.
During cross-dataset zero-shot transfer, all source-developed components, prompting templates, and model parameters remain strictly frozen. The standardized tool interfaces, the TPS schema, and procedural heuristics transfer seamlessly across benchmarks; numerical evidence is computed entirely on-the-fly for each target instance.
Target tasks steer the analytical process strictly via dynamic tool planning and runtime evidence requests, completely eliminating the need for target-side parameter training or reference-answer feedback.
Appendix~\ref{app:protocol} specifies more details of the construction and evaluation partitions.

\subsection{Perception and Tool Orchestration}
\label{sec:perception}

The perception phase grounds qualitative queries into verifiable numerical measurements via a suite of callable computational tools. Each tool $\tau$ declares its supported analyses, input requirements, and output fields. These declarations define the eligible tools $\mathcal T(X_j)$ for a given series $X_j$.
Eligibility checks account for sequence length and channel availability, while arguments specify the channel or interval to analyze. 
Conditioned on a bounded view of the series, the question, and tool descriptions, the \textsc{Perceiver} proposes an initial execution plan:
\begin{equation}
\begin{aligned}
B_j &= \bigl[(\tau_{jk},\alpha_{jk})\bigr]_{k=1}^{K_j},
\qquad \tau_{jk}\in\mathcal T(X_j),\\
\mathcal E_j &= \bigcup_{(\tau,\alpha)\in\widetilde B_j}
\operatorname{Exec}(\tau,X_j;\alpha).
\end{aligned}
\label{eq:perception-plan}
\end{equation}
where $(\tau_{jk}, \alpha_{jk})$ pairs an eligible tool with its parameterized arguments.
Memory provides procedural guidance through entry summaries and a retrieval query generated with the plan.
This proposed sequence is validated and optionally refined by the \textsc{Tool Selector} (Sec.~\ref{sec:selector}) to yield the executable plan $\widetilde B_j$.
Executing $\widetilde B_j$ against the full-resolution observations $X_j$ returns structured records $\mathcal E_j$, where each entry contains field names, values, source tools, and unique evidence identifiers. A question about a spike or level shift can pair value-profile measurements of abrupt changes with change-point localization, providing complementary evidence for TPS construction. 

The standardized execution interface accommodates tool combinations and arguments tailored to different series characteristics.
Tools operate directly on the original numerical arrays, preserving fine-grained temporal indices that can be lost in text serialization. The extracted fields retain their analytical meaning across domains, giving the \textsc{Answerer} a consistent basis for interpreting the evidence.
Appendix~\ref{app:tps-schema} details tool families, applicability checks, and an illustrative execution.

\subsection{Time-Series Perception State}
\label{sec:tps}

The TPS defines a standardized interface for reading measured series properties and adding question-specific analyses. A stable core organizes statistical and temporal fields, while named extensions retain the scope and results of additional calculations. Its state $Z_r$ is defined by
\begin{equation}
\begin{aligned}
Z_r = (\mathcal C,\mathcal D_r,\mathcal Q),\qquad
\mathcal C = \bigl\{j\mapsto(\iota_j,\mathcal S_j,\mathcal F_j,h_j)\bigr\}_{j=1}^{m},
\end{aligned}
\label{eq:tps-state}
\end{equation}
where $\mathcal C$ represents the static core perception, $\mathcal D_r$ denotes dynamic question-specific extensions, and $\mathcal Q$ denotes the set of required evidence types.
The core $\mathcal C$ provides a reusable description of the series. The question determines which measurements are acquired, while their meaning remains defined by the numerical operations. 
The dynamic extensions $\mathcal D_r$ capture analyses whose setup depends on the current task, including comparisons among supplied candidates and combinations of measured properties.
The \textsc{Answerer} interprets both under the question context and answer requirements, retaining the core as task-specific analysis expands. The core is fixed within a question, while its measured values are recomputed for each new input series. The same field definitions therefore apply to newly measured series across tasks.

Fixed rules derive $\mathcal Q$ from question text, task descriptions, and public task specifications for coverage checks during answering. An evidence type identifies a required field or analytical result, such as a trend estimate or a comparison between candidate series. These requirements remain fixed for the question while the harness acquires the corresponding evidence.

\vspace{-8pt}
\paragraph{Evidence grounding.}
Within the core $\mathcal C$, $\iota_j$ stores metadata (series identities, channel configurations, and valid temporal spans); $\mathcal S_j$ and $\mathcal F_j$ record statistical summaries (\textit{e.g.}, mean, variance, quantile bounds) and temporal primitives (\textit{e.g.}, trend monotonicity, dominant frequencies, prominent changepoints), respectively; and $h_j$ is an LLM-generated summary.
The \textsc{Perceiver} selects which measured fields to retain and generates the summary $h_j$. The runtime copies field values directly from the corresponding tool records in $\mathcal E_j$, checks their supporting references, and preserves links to their sources. Unmeasured fields remain unavailable. The resulting core preserves measured properties and their provenance for interpretation.

Each analytical extension has the form $\mathcal D_r[\nu]=(q,V,I,\delta)$, linking an analysis request $q$ to returned evidence $V$, supporting identifiers $I$, and a coverage status $\delta\in\{\text{complete},\text{incomplete}\}$. An LLM planning call uses the question, evidence requirements, and core findings to populate $\mathcal D_0$ before answering. During answering, further requests address needs that emerge when interpreting this evidence. Both stages use the same extension structure and available tools, keeping question-specific analysis compatible with the core state. In the level-shift case (Appendix~\ref{app:harness-cases}), the core records a $-2.48$ drop at index 32; an extension gathers anomaly and change-point evidence to characterize it.

\subsection{Answering with the Harness Loop}
\label{sec:answering}

The \textsc{Answerer} evaluates the tuple $(c, \kappa, Z_r)$ to determine whether to emit a terminal response or initiate a targeted evidence acquisition step in Eq.~(\ref{eq:framework-flow}).
Before evidence-conditioned answering, the same backbone generates an initial hypothesis from the series and question without TPS evidence. This provisional answer is stored in a named extension, separately from the measured properties. The \textsc{Answerer} reassesses it against the accumulated evidence; Appendix~\ref{app:harness-implementation} details the preliminary call.
The iterative loop unfolds through three consecutive phases:
\begin{enumerate}[leftmargin=*]
\vspace{-5pt}
\item \textbf{Gap inspection.} The runtime exposes missing evidence types:
\begin{equation}
\mathcal M_r=\mathcal Q\setminus\operatorname{Cov}(Z_r),
\label{eq:evidence-gap}
\end{equation}
where $\operatorname{Cov}$ denotes evidence types supplied by the available TPS fields and analytical results. Coverage tracks whether the required evidence is present; the \textsc{Answerer} assesses whether it resolves the question or motivates further analysis.
\item \textbf{Targeted re-perception.} A request $q_r=(J_r,O_r,A_r,\mathcal Q_r)$ specifies target series indices, operations, arguments, and required evidence. After validation and execution, the evidence bundle $V_r$, identifiers $I_r$, and status $\delta_r$ are appended under a fresh name $\nu_r$:
\begin{equation}
\begin{aligned}
\mathcal D_{r+1} = \mathcal D_r\cup
\{\nu_r\mapsto(q_r,V_r,I_r,\delta_r)\},\qquad
Z_{r+1} = (\mathcal C,\mathcal D_{r+1},\mathcal Q).
\end{aligned}
\label{eq:tps-update}
\end{equation}
Unsupported, duplicate, or out-of-budget requests are rejected.
\item \textbf{Constrained answer synthesis.} The \textsc{Answerer} can commit a response at any iteration and proceeds with the available state when the request budget is exhausted. The output is parsed according to $\kappa$, with an overall step limit bounding execution.
\vspace{-4pt}
\end{enumerate}
In the level-shift case, the abrupt drop identifies where the series changes, while the persistence of the new level distinguishes a level shift from an isolated spike. The \textsc{Answerer} maps this measured pattern to the requested answer option. The measurements describe the series; $(c,\kappa)$ determines their interpretation and answer form. Further requests extend $\mathcal D_r$ while retaining $\mathcal C$, allowing contextual reasoning to continue through the same numerical perception interface.

\subsection{RL-Based Tool Selection}
\label{sec:selector}

While dynamic tool invocation ensures expressive perception, executing all candidate tools within the initial tool plan incurs substantial latency and computational overhead.
We train a lightweight \textsc{Tool Selector} with final-answer feedback to refine that plan. 
Tool selection is framed as a Markov Decision Process $(\mathcal S,\mathcal A,P,r,H)$. An episode spans a complete source question. 
At decision step $t$,
\begin{equation}
\begin{aligned}
s_t = (u,z_t,\mathcal T_t,B_t),\qquad
a_t \in \mathcal A(s_t)=\{A\subseteq\mathcal T_t:A\ne\varnothing\},
\end{aligned}
\label{eq:tool-mdp}
\end{equation}
where $u$ is the question input defined in Sec.~\ref{sec:framework-overview}, $z_t$ records the current series, execution history, and remaining analysis budget, $\mathcal T_t$ contains the eligible tools, and $B_t$is the base proposal from the \textsc{Perceiver}.
Each action selects a tool subset for execution. Transition $P$ executes the selected tools, populates the harness state, and progresses either to subsequent series or to the final answering stage within a bounded horizon $H$.
The policy uses the \textsc{Perceiver}'s proposal $B_t$ as an informed prior for tool selection. For candidate tool $i \in \mathcal T_t$, we define a prior inclusion probability $p^0_{t,i}$ ($p^0_{t,i} > 0.5$ if proposed in $B_t$, and a base rate otherwise). 
A lightweight scoring network $f_\theta$ maps a dense feature vector $\phi_{t,i} = \phi(s_t, i)$ (encoding series characteristics, tool categories, and query semantics) to an additive logit shift:
\begin{equation*}
p_{t,i}=\sigma\!\left(\operatorname{logit}(p^0_{t,i})+f_\theta(\phi_{t,i})\right),
\end{equation*}
where $\sigma$ is the sigmoid function. Tool inclusion decisions follow a joint Bernoulli policy conditioned on selecting at least one operation; Appendix~\ref{app:selector} gives its normalized probability in Eq.~(\ref{eq:tool-policy}). 
Initializing $f_\theta$ near zero preserves the \textsc{Perceiver}'s initial proposal prior, allowing stable policy exploration. The default deployment rule thresholds $p_{t,i} \ge 0.5$ deterministically (defaulting to $\arg\max_i p_{t,i}$ if no tool exceeds the threshold). Conservative deployment moderates the learned adjustment and retains additional evidence through the constraints in Appendix~\ref{app:selector-deployment}.

The policy is optimized to maximize answer correctness while minimizing superfluous tool calls:
\begin{equation}
R_{\mathrm{tool}}(u)=\mathbf 1[\hat y=y]-\lambda_{\mathrm{tool}}N_{\mathrm{base}}(u),
\label{eq:tool-reward}
\end{equation}
where $\hat y$ is the predicted answer, $y$ is the source reference answer, and $N_{\mathrm{base}}(u)$ counts executions governed by the selector across the question. The coefficient $\lambda_{\mathrm{tool}} > 0$ sets the execution penalty, using tool count as a cost proxy for numerical analysis overhead.
We optimize $f_\theta$ using Proximal Policy Optimization (PPO)~\citep{ppo}. The policy gradient is computed over multiple trajectory rollouts per source instance, with the baseline defined as the leave-one-out rollout mean to reduce variance.
Once trained on source tasks, the parameters $\theta$ of the \textsc{Tool Selector} are frozen alongside the backbone LLM, delivering an adaptive, cost-effective perception frontend across all zero-shot target evaluations.

\section{Experiments}
\label{sec:experiments}

\subsection{Experimental setup}
We evaluate on TSQA tasks from six benchmarks: TSRBench~\citep{tsrbench}, Time-MQA~\citep{timemqa}, ChatTime~\citep{chattime}, CaTSBench~\citep{catsbench}, TSAQA~\citep{tsaqa}, and TimeOmni-1~\citep{timeomni}. These benchmarks span multiple question types and assess a broad range of time-series understanding and reasoning capabilities, with diverse time-series inputs, question contexts, and answer requirements. TSRBench serves as the source benchmark for \fullname{} construction, while the other five provide targets for cross-dataset evaluation.
Appendix~\ref{app:protocol} provides further details.

We evaluate our approach in two settings:
$\bullet$ \emph{In-distribution (ID) evaluation.}  We use TSRBench to develop the harness, including constructing numerical tools and organizing analysis knowledge into Memory. Following prior work~\citep{timeclaw}, we use the $10\%$ TSRBench test split for evaluation. 
$\bullet$ \emph{Zero-shot evaluation.} We directly deploy the TSRBench-developed harness on the five target benchmarks under a zero-shot protocol. For comparisons among agentic methods, we randomly sample a fixed subset of $412$ QA examples per target benchmark, covering different question types. This setting tests transfer across series, questions, and answer requirements without target-side training, tuning, or reference-answer feedback.

We compare \fullname{} with three groups of baselines. $\bullet$ Task-specialized models include the released ChatTime~\citep{chattime} and Time-MQA checkpoints~\citep{timemqa}. The zero-shot baseline directly answers each question from the input time series and accompanying context, without access to the harness tools or the memory. $\bullet$ To assess \fullname{} against agentic approaches, we include chain-of-thought (CoT)~\citep{cot}, ReAct~\citep{react}, and the released TimeClaw framework~\citep{timeclaw}. $\bullet$ We evaluate GPT-5.6-sol and the open-source Qwen3.8-27B. We implement CoT, ReAct, and TimeClaw, using Qwen3.8-27B as the backbone. 

We report accuracy (\%) across each benchmark, computed as the proportion of questions correctly answered under its native parsing and scoring protocols. Model responses that are invalid, unparseable, or format-violating are strictly penalized as incorrect. 
Appendix~\ref{app:metrics} provides full details on answer parsing rules and metric accounting.

\subsection{Main Results}

\begin{table}[t]
\vspace{-5pt}
\centering
\caption{\textbf{Cross-dataset zero-shot TSQA evaluation} (Accuracy, \%).
\fullname{} is developed exclusively on the source benchmark (TSRBench) and deployed for zero-shot evaluation on five unseen target benchmarks.
\fullname{}$^{\star}$ incorporates the learned RL-based \textsc{Tool Selector}.
Shaded orange cells indicate \textcolor{orange!65!black}{\textbf{in-domain}} training for ChatTime and Time-MQA, directly contrasting with the strictly zero-shot deployment of \modelname{}.
The best results within each backbone group are highlighted in \textbf{bold}; second-best results within the open-weight agentic group are \underline{underlined}.}
\vspace{3pt}
\label{tab:transfer}
\begingroup
\setlength{\tabcolsep}{3pt}
\renewcommand{\arraystretch}{1.18}
\newcommand{\transferbest}[1]{\textbf{#1}}
\newcommand{\transfersource}[1]{\begingroup\setlength{\fboxsep}{1.5pt}\colorbox{orange!15}{#1}\endgroup}
\newcommand{\transfergroup}[2]{\multicolumn{8}{c}{\textbf{\textit{#2}}}\\}
\newcommand{\transfertabular}{%
\begin{tabular}{@{}ll*{6}{c}@{}}
\toprule
\multirow{2}{*}{\textbf{Method}} & \multirow{2}{*}{\textbf{Base LLM}} & \multirow{2}{*}{\textbf{TSRBench}} & \multicolumn{5}{c}{\textbf{Target Benchmarks}} \\
\cmidrule(lr){4-8}
 & & & \textbf{Time-MQA} & \textbf{ChatTime} & \textbf{CaTSBench} & \textbf{TSAQA} & \textbf{TimeOmni} \\
\midrule
\transfergroup{black!7}{Specialized Time-Series Models}
\addlinespace[2pt]
ChatTime & LLaMA-2-7B & 19.95 & 28.79 & \transfersource{88.77} & 9.63 & 11.36 & 19.66 \\
\multirow[t]{3}{*}{Time-MQA} & Qwen2.5-7B & 30.47 & \transfersource{54.98} & 37.07 & 26.05 & 45.07 & 32.04 \\
 & Llama3-8B & 20.95 & \transfersource{54.58} & 33.34 & 51.27 & 40.30 & 33.25 \\
 & Mistral-7B & 36.27 & \transfersource{55.91} & 35.47 & 47.79 & 42.61 & 31.55 \\
\midrule
\transfergroup{black!7}{Proprietary Model}
\addlinespace[2pt]
Zero-shot & \multirow{2}{*}{GPT-5.6-sol} & 66.75 & 61.65 & 89.73 & 98.79 & 57.77 & \transferbest{75.73} \\
\textbf{\modelname{}} &  & \transferbest{72.09} & \transferbest{67.48} & \transferbest{96.88} & \transferbest{99.27} & \transferbest{58.50} & 74.76 \\
\midrule
\transfergroup{black!7}{Open-Weight Model and Agentic Methods}
\addlinespace[2pt]
Zero-shot & \multirow{6}{*}{Qwen3.8-27B} & 43.69 & 58.01 & 74.13 & 92.96 & 52.18 & 50.97 \\
CoT &  & 37.14 & 61.65 & 74.03 & 93.69 & 51.94 & 50.97 \\
ReAct &  & 37.62 & \underline{61.89} & 77.67 & 91.50 & 46.12 & 42.23 \\
TimeClaw &  & \underline{49.51} & 54.37 & 80.10 & \transferbest{96.84} & 48.30 & \underline{54.61} \\
\textbf{\modelname{}} &  & \transferbest{62.62} & \transferbest{65.78} & \underline{96.10} & \underline{96.60} & \underline{59.47} & \transferbest{55.10} \\
\textbf{\modelname{}}$^{\star}$ &  & \transferbest{62.62} &  \transferbest{65.78}& \transferbest{96.12} & \transferbest{96.84} & \transferbest{59.71} & \transferbest{55.10} \\
\bottomrule
\end{tabular}}
\sbox0{\transfertabular}
\edef\transferwidth{\the\wd0}
\renewcommand{\transfergroup}[2]{\multicolumn{8}{@{}c@{}}{\begingroup\setlength{\fboxsep}{0pt}\colorbox{#1}{\makebox[\transferwidth][c]{\rule[-0.7ex]{0pt}{2.8ex}\textbf{\textit{#2}}}}\endgroup}\\}
\resizebox{\textwidth}{!}{\transfertabular}
\endgroup
\vspace{-8pt}
\end{table}

\fullname{} improves source-benchmark accuracy with both open-weight and proprietary backbones (Table~\ref{tab:transfer}). On TSRBench, accuracy increases from $43.69\%$ to $62.62\%$ with Qwen3.8-27B and from $66.75\%$ to $72.09\%$ with GPT-5.6-sol.
The source-developed harness also transfers across datasets without target-side training or answer feedback. With Qwen3.8-27B, \fullname{} improves over zero-shot answering on all five target benchmarks. The largest gain occurs on ChatTime, where accuracy rises from $74.13\%$ to $96.10\%$. Gains on Time-MQA, CaTSBench, TSAQA, and TimeOmni are $7.77$, $3.64$, $7.28$, and $4.13$ percentage points, respectively. With GPT-5.6-sol, \fullname{} improves accuracy on four of five targets, including increases from $61.65\%$ to $67.48\%$ on Time-MQA and from $89.73\%$ to $96.88\%$ on ChatTime. TimeOmni is the exception, with accuracy decreasing from $75.73\%$ to $74.76\%$. Transfer benefits therefore extend across both backbones, although their magnitude and direction depend on the target benchmark.

Among methods using Qwen3.8-27B, \fullname{} achieves the highest accuracy on TSRBench and four of the five target benchmarks. On TSRBench, it exceeds TimeClaw, the strongest baseline, by $13.11$ percentage points. On Time-MQA, ChatTime, TSAQA, and TimeOmni, it exceeds the strongest baseline by $3.88$, $16.00$, $7.28$, and $0.49$ percentage points, respectively. On CaTSBench, \fullname{} reaches $96.60\%$, compared with $96.84\%$ for TimeClaw. These comparisons support the effectiveness of \fullname{} across diverse TSQA tasks under a shared backbone. The Tool Selector variant, \fullname{}$^{\star}$, is reported separately; Sec.~\ref{sec:tool-selection-results} analyzes its execution cost and transfer results.

\subsection{Ablation Study of Tools and Memory}

In Table~\ref{tab:ablation}, we isolate the individual contributions of numerical tools and procedural memory within \fullname{} across both in-domain and zero-shot transfer settings.
On the source benchmark (TSRBench), both components yield substantial performance gains. Disabling numerical tools precipitates a severe performance drop of $16.50$ percentage points ($62.62\% \to 46.12\%$), demonstrating that raw LLM reasoning alone cannot reliably parse intricate numerical patterns. Removing both components lowers accuracy to $42.48\%$.
In cross-dataset zero-shot evaluation, Tool consistently improves accuracy across all five target benchmarks. Removing Tool lowers average target accuracy from $74.61\%$ to $63.06\%$. Memory provides limited gains in this setting: average target accuracy is $74.76\%$ without Memory, compared with $74.61\%$ for the full harness. The full harness exceeds the variant without either component by $11.21$ points in average target accuracy. These results establish Tool as a key contributor to cross-dataset transfer, while Memory is most beneficial within its construction domain.

\begin{table}[t]
\vspace{-10pt}
\centering\small
\caption{\textbf{Ablation study of core components across in-domain and zero-shot target datasets} (Accuracy, \%). We evaluate variants of \modelname{} by disabling numerical tools and procedural memory using Qwen3.8-27B as the backbone LLM. Numerical tool orchestration proves indispensable for both in-domain and transfer accuracy. Best results are shown in \textbf{bold}.}
\label{tab:ablation}
\vspace{3pt}
\setlength{\tabcolsep}{6.5pt}
\begin{tabular}{lcccccc}
\toprule
\multirow{2}{*}{Variant} & In-domain & \multicolumn{5}{c}{Zero-shot} \\
\cmidrule(lr){2-2}\cmidrule(lr){3-7}
 & TSRBench & Time-MQA & ChatTime & CaTSBench & TSAQA & TimeOmni \\
\midrule
Full \modelname{} & \textbf{62.62} & 65.78 & 96.10 & \textbf{96.60} & \textbf{59.47} & \textbf{55.10} \\
w/o Tool & 46.12 & 60.92 & 63.59 & 91.02 & 47.33 & 52.43 \\
w/o Memory & 60.44 & \textbf{68.45} & \textbf{96.12} & 96.12 & 58.74 & 54.37 \\
w/o Tool + Memory & 42.48 & 60.92 & 63.35 & 91.02 & 48.79 & 52.91 \\
\bottomrule\end{tabular}
\end{table}

\begin{figure}[t]
\centering
\includegraphics[width=\textwidth]{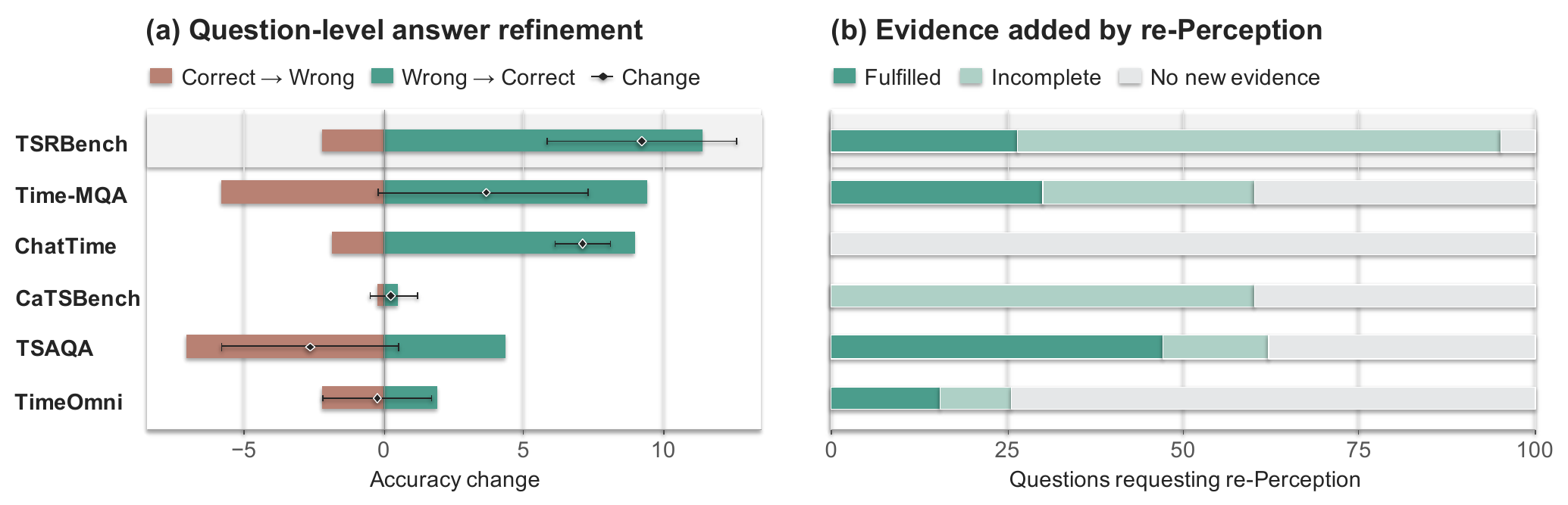}
\vspace{-22pt}
\caption{\textbf{Analysis of answer refinement and active re-perception dynamics.}
(a) Transitions from initial pre-TPS hypotheses to final verified answers across benchmarks. Diamond markers represent net accuracy changes in percentage points.
(b) Evidence fulfillment rates among queries triggering the re-perception loop (\%). \emph{Fulfilled} and \emph{Incomplete} indicate acquired evidence with full and partial request coverage, respectively. Gray indicates that no additional evidence was obtained through re-perception.}
\label{fig:pipeline-revisions}
\vspace{-10pt}
\end{figure}

\subsection{Answer Refinement and Re-Perception}
\label{sec:pipeline-analysis}

We analyze how the decoupled workflow refines initial reasoning and acquires incremental evidence, using GPT-5.6-sol rollouts from Table~\ref{tab:transfer}.
Results are shown in Figure~\ref{fig:pipeline-revisions}.

\vspace{-8pt}
\paragraph{Question-level answer refinement.}
Figure~\ref{fig:pipeline-revisions}(a) illustrates the transitions from ungrounded pre-TPS hypotheses to final grounded predictions. Across the majority of benchmarks, error corrections substantially surpass regressions, yielding significant net accuracy gains of $+9.22$ percentage points on TSRBench and $+7.10$ on ChatTime. 
Time-MQA shows the same positive trend, yielding a net accuracy gain of $3.64$ points.
While TSAQA and TimeOmni exhibit minor declines of $-2.67$ and $-0.24$ points, the overall results show that numerical perception and TPS-grounded reasoning improve initial judgments across several benchmarks, although the benefit varies across tasks.

\vspace{-8pt}
\paragraph{Evidence added by re-perception.}
When the \textsc{Answerer} detects that existing state observations are insufficient to resolve ambiguous temporal queries, the harness activates targeted re-perception. As shown in Figure~\ref{fig:pipeline-revisions}(b), the active feedback loop successfully acquires and fulfills missing evidence for $95.2\%$ of requesting queries on TSRBench, $62.1\%$ on TSAQA, and $25.5\%$ on TimeOmni.
Notably, ChatTime triggers zero re-perception requests; rather than a limitation, this indicates the cost-conscious selectivity of our harness, as the base perception state already provides sufficient quantitative support for its conversational queries.
The harness loop thus functions as an adaptive, on-demand analytical mechanism that selectively expends computational budget only when resolving genuine evidentiary deficits. Detailed transition matrices and qualitative execution cases are provided in Appendix~\ref{app:answer-revisions} and Appendix~\ref{app:harness-cases}.

\subsection{Analyses on Learned Tool Selection}
\label{sec:tool-selection-results}
The learned \textsc{Tool Selector} enables \fullname{} to dynamically steer evidence acquisition across distinct accuracy and efficiency regimes at test time without retraining the policy in the target domain. 

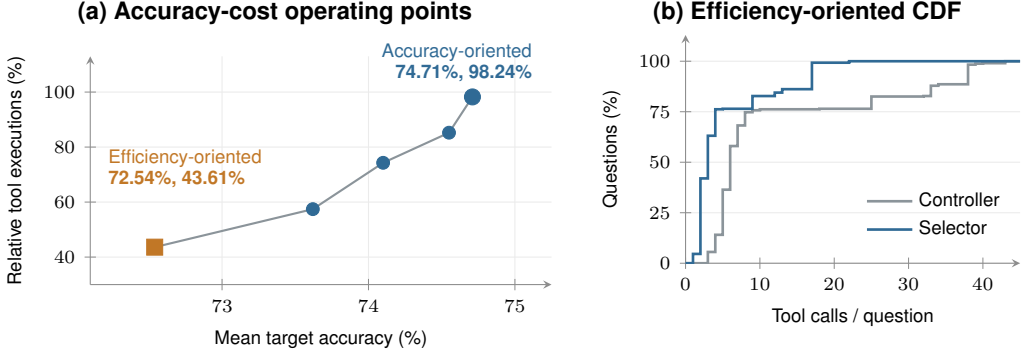
\begin{figure}[t]
\centering
\begingroup
\definecolor{selectblue}{RGB}{48,106,149}
\definecolor{selectamber}{RGB}{193,119,38}
\definecolor{selectgray}{RGB}{139,148,154}
\def\effAccuracy{72.54}
\def\accAccuracy{74.71}
\def\effCost{43.61}
\def\accCost{98.24}
\begin{minipage}[t]{0.55\textwidth}
\centering
{\sffamily\footnotesize\bfseries (a) Accuracy-cost operating points}\par\vspace{3pt}
\begin{tikzpicture}
\begin{axis}[
width=\linewidth,height=4.6cm,
scale only axis=false,
xmin=72.1,xmax=75.25,ymin=30,ymax=113,
xtick={73,74,75},ytick={40,60,80,100},
xlabel={Mean target accuracy (\%)},
ylabel={Relative tool executions (\%)},
axis lines=left,axis line style={black!45},tick style={black!45},
tick label style={font=\sffamily\scriptsize},
label style={font=\sffamily\scriptsize},
grid=major,grid style={black!8},
clip=false]
\addplot[selectgray,line width=0.8pt,no marks]
coordinates {(\effAccuracy,\effCost) (73.62,57.43) (74.10,74.28) (74.55,85.23) (\accAccuracy,\accCost)};
\addplot[only marks,mark=*,mark size=2.4pt,selectblue]
coordinates {(73.62,57.43) (74.10,74.28) (74.55,85.23)};
\addplot[only marks,mark=square*,mark size=3pt,selectamber]
coordinates {(\effAccuracy,\effCost)};
\addplot[only marks,mark=*,mark size=3pt,selectblue]
coordinates {(\accAccuracy,\accCost)};
\node[anchor=south west,font=\sffamily\scriptsize,text=selectamber,align=left]
at (axis cs:72.16,62) {Efficiency-oriented\\\textbf{72.54\%, 43.61\%}};
\node[anchor=south east,font=\sffamily\scriptsize,text=selectblue,align=right]
at (axis cs:75.18,100.5) {Accuracy-oriented\\\textbf{74.71\%, 98.24\%}};
\end{axis}
\end{tikzpicture}
\end{minipage}\hfill
\begin{minipage}[t]{0.43\textwidth}
\centering
{\sffamily\footnotesize\bfseries (b) Efficiency-oriented CDF}\par\vspace{3pt}
\begin{tikzpicture}
\begin{axis}[
width=\linewidth,height=4.6cm,
xmin=0,xmax=45,ymin=0,ymax=113,
xtick={0,10,20,30,40},ytick={0,25,50,75,100},
xlabel={Tool calls / question},ylabel={Questions (\%)},
axis lines=left,axis line style={black!45},tick style={black!45},
tick label style={font=\sffamily\scriptsize},
label style={font=\sffamily\scriptsize},
ymajorgrids=true,grid style={black!8},
legend style={font=\sffamily\scriptsize,draw=none,fill=none,at={(0.98,0.06)},anchor=south east},
legend cell align=left]
\addplot[const plot,selectgray,line width=1pt,no marks]
coordinates {(0,0.00000000) (3,5.58252427) (4,14.07766990) (5,36.40776699) (6,58.00970874) (7,68.20388350) (8,74.75728155) (9,75.72815534) (10,76.21359223) (18,76.45631068) (25,82.52427184) (32,82.76699029) (33,87.86407767) (34,88.59223301) (38,98.30097087) (39,98.78640777) (40,99.02912621) (43,100.00000000) (45,100.00000000)};
\addlegendentry{Controller}
\addplot[const plot,selectblue,line width=1pt,no marks]
coordinates {(0,0.00000000) (1,4.61165049) (2,41.99029126) (3,63.10679612) (4,76.21359223) (5,76.45631068) (9,82.76699029) (12,84.46601942) (13,86.16504854) (17,99.27184466) (22,100.00000000) (45,100.00000000)};
\addlegendentry{Selector}
\end{axis}
\end{tikzpicture}
\end{minipage}
\endgroup
\vspace{-8pt}
\caption{\textbf{Deployment choices for learned tool selection.} (a) Mean accuracy and total tool-execution ratios across all five target benchmarks (matched Controller = 100\%). All five points are measured configurations; lines connect adjacent points. (b) The separate TSRBench cumulative distribution under efficiency-oriented deployment.}
\label{fig:tool-efficiency}
\vspace{-5pt}
\end{figure}

As shown in Figure~\ref{fig:tool-efficiency}(a), empirical evaluations across five operating configurations demonstrate this flexible control across all target benchmarks.
Under the accuracy-oriented deployment (\fullname{}$^{\star}$), the harness attains $74.71\%$ macro-average target accuracy while reducing tool executions by $1.76\%$ relative to the validated proposal baseline.
Under the efficiency-oriented regime, the selector slashes overall tool executions by $\bf 56.39\%$ while incurring only a modest $2.17\%$ decrease in accuracy ($72.54\%$).
Throughout these comparisons, all tool costs are normalized to the matched Controller (the \textsc{Perceiver}'s validated plan), with downstream evidence processing fixed within each comparison. 
These trajectories highlight that the \textsc{Tool Selector} provides effective, fine-grained budget control, illustrating that aggressive computational savings can be achieved with minimal degradation in reasoning fidelity.

Furthermore, distributional analysis on the source benchmark (Figure~\ref{fig:tool-efficiency}(b)) reveals that efficiency-oriented selection substantially suppresses the heavy upper tail of tool invocations. By penalizing redundant queries on complex, multi-series inputs, the learned policy effectively mitigates worst-case execution latency and bounds peak resource consumption.
Together, learned preferences and evidence-preservation constraints equip \fullname{} with configurable, cost-predictable perceptual efficiency across deployment environments. Details are documented in Appendices~\ref{app:selector-deployment} and~\ref{app:selector-operating-points}.

\section{Conclusion and Limitations}

We presented \fullname{}, an agentic framework that makes source-developed analytical capabilities reusable across TSQA datasets. 
To overcome the dual transfer bottlenecks caused by coupled series and task shifts, our core design decouples continuous numerical perception from context-specific reasoning via an intermediate Time-Series Perception State (TPS). Within this framework, collaborative \textsc{Perceiver} and \textsc{Answerer} roles coordinate tool execution, procedural memory, and active evidence-request loops, while a reinforcement-learning \textsc{Tool Selector} adaptively optimizes the trade-off between answer accuracy and tool execution budget.
Extensive evaluations demonstrate that an analytical workflow constructed exclusively on a single source dataset reliably generalizes across five unseen target benchmarks with diverse data distributions and answer specifications.

Despite its strong transferability, \fullname{} has three main limitations. First, its perceptual capabilities are constrained by the pre-registered operations in its tool suite.
Second, our \textsc{Tool Selector} uses a frozen policy network during zero-shot deployment, leaving test-time adaptation under severe domain shifts unexplored.
Finally, zero-shot performance inherently depends on the diversity and task overlap of the source benchmark, highlighting the need for larger, open-world TSQA datasets to construct the harness system and evaluate broader generalizability.

\bibliography{references}
\bibliographystyle{iclr2027_conference}

\appendix
\clearpage
\section{Datasets and Evaluation Details}
\label{app:protocol}

This section supplements the experimental setup in Sec.~\ref{sec:experiments}, specifying the QA tasks, input representation, evaluation subsets, and answer scoring.

\subsection{Benchmark coverage}

Table~\ref{tab:appendix-benchmarks} summarizes the evaluated QA tasks. TSRBench~\citep{tsrbench} supplies the construction data and in-domain evaluation. ChatTime~\citep{chattime}, Time-MQA~\citep{timemqa}, CaTSBench~\citep{catsbench}, TSAQA~\citep{tsaqa}, and TimeOmni-1~\citep{timeomni} test transfer across series, contexts, and answer requirements.

\begin{table}[ht]
\centering\small
\caption{Time-series inputs, QA capabilities, and answer forms covered by the evaluation.}
\label{tab:appendix-benchmarks}
\begingroup
\setlength{\tabcolsep}{3pt}
\renewcommand{\arraystretch}{1.15}
\newcommand{\benchcell}[1]{{\raggedright #1\par}}
\begin{tabular}{@{}p{0.16\textwidth}p{0.25\textwidth}p{0.35\textwidth}p{0.18\textwidth}@{}}
\toprule
\benchcell{Benchmark} & \benchcell{Time series and context} & \benchcell{QA capabilities} & \benchcell{Answer forms} \\[3pt]
\midrule
\benchcell{TSRBench} & \benchcell{Single or multiple series with domain context} & \benchcell{Perception, reasoning, prediction, and decision-making} & \benchcell{Multiple choice} \\[3pt]
\benchcell{ChatTime} & \benchcell{Synthetic series of varying lengths} & \benchcell{Trend, volatility, seasonality, and outlier recognition} & \benchcell{Multiple choice} \\[3pt]
\benchcell{Time-MQA} & \benchcell{Real-world series with contextual descriptions} & \benchcell{Statistical properties, patterns, anomalies, and structural changes} & \benchcell{True/false; multiple choice} \\[3pt]
\benchcell{CaTSBench} & \benchcell{Series paired with textual descriptions} & \benchcell{Caption and series retrieval, description consistency, and pairwise comparison} & \benchcell{True/false; multiple choice} \\[3pt]
\benchcell{TSAQA} & \benchcell{Cross-domain series and temporal patches} & \benchcell{Classification, anomalies, characterization, comparison, transformations, and temporal relationships} & \benchcell{True/false; multiple choice; ordered labels} \\[3pt]
\benchcell{TimeOmni-1} & \benchcell{Series with event and domain context} & \benchcell{Causality discovery, scenario understanding, and decision-making} & \benchcell{Multiple choice} \\[3pt]
\bottomrule
\end{tabular}
\endgroup
\end{table}

\subsection{Input representation and transfer protocol}

We use the input notation $u=(X,c,\kappa)$ introduced in Sec.~\ref{sec:method}:
\begin{itemize}
\item \textbf{Observations $X$.} Dataset adapters separate numerical observations from the question, retaining the correspondence between each series and its description. Tools operate on these observations; the Controller uses the bounded view described in Sec.~\ref{sec:method}.
\item \textbf{Context $c$.} The question, domain descriptions, and answer choices remain available for planning and contextual reasoning. Retrieval and comparison questions retain their candidate series or descriptions; temporal-ordering questions retain their patch identities.
\item \textbf{Requirements $\kappa$.} The requested answer form determines output parsing. Reference answers provide source feedback and determine evaluation scores.
\end{itemize}

TSRBench construction examples provide feedback for developing the tools and organizing analysis knowledge into Memory. In-domain evaluation uses the held-out approximately 10\% TSRBench subset following the TimeClaw protocol~\citep{timeclaw}. For each backbone, the entire system is fixed before target evaluation, including model weights, all prompts, tools and activation rules, Memory and retrieval rules, the Tool Selector when used, dataset adapters, evidence-requirement mappings, and answer parsers. The analysis remains adaptive at inference through tool plans and evidence requests, as specified in Eq.~(\ref{eq:framework-flow}). The independent zero-shot baseline receives the series and question directly, without Tool or Memory access.

\paragraph{Source-data roles.}
Within TSRBench, construction examples supply execution traces and answer feedback for developing Tool and Memory resources. The Tool Selector uses a source-training subset for policy updates and a separate source-development subset for checkpoint selection. Its configuration is given in Appendix~\ref{app:selector}; the resource-construction procedure is detailed in Appendix~\ref{app:construction}. The held-out TSRBench test partition is excluded from resource construction, policy training, and checkpoint selection. Target benchmarks are excluded from system development and tuning; their reference answers are used only to score completed predictions. This separation prevents evaluation data from entering system development. The TSRBench test partition measures ID performance, while the other five benchmarks assess transfer of the source-developed system.

\paragraph{Inference configuration.}
API-based LLM components use official provider APIs. The TSRBench harness runs use temperature 0 and a maximum of 768 generated tokens per LLM call for both backbones. Planning, TPS synthesis, and answering use separate role-specific calls to the same backbone. Structured responses receive one additional parsing attempt when needed. Tool-plan and evidence-request budgets are specified in Appendix~\ref{app:runtime-limits}.

\subsection{Evaluation subsets}

For cross-evaluation, we use the full test sets provided by prior work. For comparisons among agentic methods, we randomly sample a fixed subset of $412$ questions per benchmark, covering different question types, and evaluate the compared methods on the same subset. The sampled subsets for agentic comparisons on the target benchmarks are described below.
\begin{itemize}
\item \textbf{ChatTime.} We randomly sample $412$ questions covering four task categories: Trend, Seasonality, Volatility, and Outliers.
\item \textbf{Time-MQA.} We randomly sample $412$ questions from the objective test set, covering different question types.
\item \textbf{CaTSBench and TSAQA.} Each benchmark uses a fixed subset of $412$ randomly sampled questions covering different task categories. CaTSBench plot-retrieval questions requiring unavailable candidate images are excluded from the text-based evaluation; numerical-series and text-description tasks are retained.
\item \textbf{TimeOmni-1.} We randomly sample $412$ accuracy-scored questions from the OOD split, covering causality discovery, scenario understanding, and decision-making. All of these questions are target data relative to TSRBench construction.
\end{itemize}
Comparisons among agentic methods on TSRBench use the same held-out $412$-question test subset.

\subsection{Answer parsing and scoring}
\label{app:metrics}

Let $P_d$ denote the answer parser for benchmark $d$, $\hat y_i$ the generated response, and $y_i$ the reference. Accuracy is
\begin{equation}
\operatorname{Acc}_d
=\frac{100}{N_d}\sum_{i=1}^{N_d}
\mathbf{1}\!\left[P_d(\hat y_i)=P_d(y_i)\right],
\label{eq:appendix-accuracy}
\end{equation}
where $N_d$ is the number of completed inference runs included in scoring. An invalid or unparseable final response from a completed run contributes zero.
\begin{itemize}
\item \textbf{Multiple choice.} Responses are mapped to an option label using the benchmark's answer format and compared with the reference label.
\item \textbf{True/false.} Boolean answers are normalized before exact comparison.
\item \textbf{Temporal ordering.} Exact agreement of the complete ordered answer with the reference is required after text normalization.
\end{itemize}
When reporting an average over target benchmarks, we use the unweighted mean $\overline{\operatorname{Acc}}=\frac{1}{5}\sum_{d\in\mathcal D_{\mathrm{target}}}\operatorname{Acc}_d$, excluding TSRBench. Percentage-point differences are computed from unrounded accuracies.

\section{Harness Implementation Details}
\label{app:tps-schema}
\label{app:harness-implementation}

This section supplements Section~\ref{sec:method} with the tool interfaces, construction of the intermediate TPS, and evidence updates underlying the decoupling of perception and answering. Two successful target-benchmark executions illustrate how numerical findings support contextual judgments.

\subsection{Source construction and frozen deployment}
\label{app:construction}

The development stage in Figure~\ref{fig:method} converts TSRBench execution feedback into the Tool and Memory resources used by the harness. Its data roles are specified in Appendix~\ref{app:protocol}. Construction proceeds in batches of 16 source questions:
\begin{enumerate}
\item \textbf{Run and inspect.} The current harness processes each question and records its tool evidence, TPS, preliminary hypothesis, and final prediction. Reference answers are then used to assess the completed batch and distinguish corrected, regressed, and persistently incorrect judgments.
\item \textbf{Develop the tool configuration.} The initial library combines general statistical and temporal operations with implemented task-specific calculations, such as signal analysis, formula evaluation, and candidate-trajectory comparison. Source development includes implementing new operations, configuring existing tools, and establishing tool--task activation rules. Source feedback determines the available specialized tool--task pairings and motivates additional diagnostic operations. For calculation-based pairings, the activation rule requires at least two supporting observations and 95\% agreement with source references. Diagnostic operations that return measurements or interpretations are enabled through checks on source examples; their outputs enter TPS for the Answerer to interpret.
\item \textbf{Organize analytical knowledge.} An LLM editor turns the batch's evidence and outcomes into Memory entries containing a general procedure, applicability conditions, recommended tools, and evidence requirements. Entries are organized by task or analytical topic. An update revises the corresponding entry when that topic already exists. Checks validate the topic and tool references and filter literal option-answer statements; task-specific checks assess proposed rules against the available source evidence.
\item \textbf{Freeze the resources.} The resulting Memory bank and enabled-tool configuration are saved for deployment. The Tool Selector is subsequently trained with these resources fixed, as detailed in Appendix~\ref{app:selector}. Target inference loads the source-developed resources and computes measurements for each new question through the same tool interfaces and TPS schema.
\end{enumerate}

\subsection{Tools and analytical guidance}

\paragraph{Tool interfaces and procedural guidance.}
The library covers statistics, trends, seasonality, volatility, change points, and cross-series relationships. Eligibility checks use declared requirements, including minimum sequence length and the availability of multiple channels. Plans specify tool names and supported arguments, which can identify a channel or temporal interval. The runtime checks names, argument types, and declared ranges before execution. Memory entries describe applicability, recommended operations, and evidence requirements. Their summaries inform planning, and a query generated with the plan retrieves relevant procedures under fixed retrieval rules.

Table~\ref{tab:harness-tools} summarizes representative numerical operations and their outputs. Each registered tool also declares supported fields, an argument schema, and applicability conditions. For example, trend estimation requires at least two observations, while cross-series analysis requires multiple channels and at least four observations. Validated arguments can adjust detection thresholds or the maximum lag. The Controller's plan is deduplicated and restricted to eligible operations before execution.

\begin{table}[ht]
\centering\small
\caption{Representative tools used to construct TPS fields. Measurements and temporal locations are computed from the input observations.}
\label{tab:harness-tools}
\begingroup
\setlength{\tabcolsep}{3pt}
\renewcommand{\arraystretch}{1.12}
\newcommand{\toolcell}[1]{{\raggedright #1\par}}
\begin{tabular}{@{}p{0.20\linewidth}p{0.43\linewidth}p{0.32\linewidth}@{}}
\toprule
\toolcell{Tool family} & \toolcell{Computation} & \toolcell{Returned information} \\[3pt]
\midrule
\toolcell{Statistics and value profile} & \toolcell{Summary statistics, quantiles, extrema, and adjacent differences} & \toolcell{Values, extrema locations, largest rise and drop} \\[3pt]
\toolcell{Trend} & \toolcell{Linear fit on normalized time; slope scaled by series standard deviation} & \toolcell{Direction, strength, normalized slope} \\[3pt]
\toolcell{Seasonality} & \toolcell{Autocorrelation and spectral analysis} & \toolcell{Period and periodicity strength} \\[3pt]
\toolcell{Volatility} & \toolcell{Early/late standard deviations and their ratio} & \toolcell{Variability level and change} \\[3pt]
\toolcell{Anomalies and change points} & \toolcell{Median-deviation and local-residual scores; mean and variance changes across windows} & \toolcell{Event positions, types, directions, and scores} \\[3pt]
\toolcell{Cross-series} & \toolcell{Contemporaneous and lagged correlations} & \toolcell{Channel pairs, correlations, and strongest lag} \\[3pt]
\bottomrule
\end{tabular}
\endgroup
\end{table}

\paragraph{Memory contents.}
An entry stores analysis guidance together with its applicability, triggers, recommended tools, and required evidence. Retrieval matches the query against the entry's content, tags, and triggers. For example, one source-developed entry explains that comparing the innovation scales of two AR(1) series requires fitting each autoregression and comparing residual standard deviations. It recommends the corresponding signal-analysis tool and identifies residual scale and autoregressive coefficient as the required evidence. This makes the relevant analytical steps available to the Controller while the tools compute their values for the current input.

\subsection{From tool outputs to TPS}

Table~\ref{tab:tps-fields} expands the statistical fields, temporal fields, and named extensions defined in Eq.~(\ref{eq:tps-state}). Field values may be scalars or structured outputs; supporting evidence identifiers link them to the producing tool records. Series identity, channels, and length remain available alongside the fields, and the LLM-generated summary provides a concise textual view.

\begin{table}[ht]
\centering\small
\caption{TPS field groups and their analytical roles.}
\label{tab:tps-fields}
\begin{tabular}{@{}p{0.23\linewidth}p{0.72\linewidth}@{}}
\toprule
Component & Contents and role \\
\midrule
Statistics $\mathcal S_j$ & Distributions and value profiles, including means, standard deviations, extrema, and their locations. \\
Temporal features $\mathcal F_j$ & Trends, periodicity, volatility, segments, changes, anomalies, and cross-series relationships. \\
Task extensions $\mathcal D_r$ & Named analyses that add evidence required by the question beyond the common fields. \\
\bottomrule
\end{tabular}
\end{table}

\paragraph{Building the intermediate representation.}
Perception uses two role-specific LLM calls: the first proposes tools and a Memory query; the second receives the resulting evidence, selects measured fields, and writes the summary $h_j$. TPS construction then proceeds as follows:
\begin{itemize}
\item \textbf{Bind values to measurements.} Each selected field is populated directly from its tool record. A value can be a scalar, a structured property such as trend direction and strength, or a list of events with their temporal positions.
\item \textbf{Retain analytical meaning and sources.} Each field stores its supporting evidence identifiers, the producing tool and version, and available confidence information. The series identity and original indices connect temporal findings back to the observations.
\item \textbf{Represent task-specific analysis.} Named extensions store the requested operations, returned evidence, and coverage status in the form $\mathcal D_r[\nu]=(q,V,I,\delta)$ from Section~\ref{sec:tps}. These can include cross-series comparisons, candidate trajectory measurements, or temporal-ordering scores.
\end{itemize}
Thus, the Answerer receives an interpretable description of what was measured and where, together with question-specific evidence. It uses the question and answer requirements to turn this description into a contextual judgment.

\paragraph{Initial hypothesis and task extensions.}
Before evidence-conditioned answering, a separate call to the same backbone receives the raw numerical series, question context, answer choices, and output requirements, without TPS evidence or reference answers. Its parsed candidate answer and response are stored in a named extension as a provisional hypothesis, separate from the measured core properties. The Answerer reassesses this hypothesis against the acquired evidence. An initial extension-planning call uses the core findings, question, evidence requirements, and available tools and Memory to propose additional analyses; the harness validates and executes these requests to populate $\mathcal D_0$.

\paragraph{Role-specific LLM interfaces.}
Table~\ref{tab:harness-role-interfaces} summarizes the information passed between the LLM roles using the same backbone. Memory summaries inform the tool plan, and retrieved entries guide the interpretation of measured evidence. The LLM selects fields and writes the summary; the harness populates their numerical values directly from tool records. The Answerer then interprets TPS under the question's context and output requirements.

\begin{table}[htbp]
\centering\small
\caption{Inputs and outputs of the LLM roles connecting Perception, TPS, and the Answerer. Tool execution and numerical field assignment are performed by the harness between calls.}
\label{tab:harness-role-interfaces}
\begingroup
\setlength{\tabcolsep}{3pt}
\renewcommand{\arraystretch}{1.12}
\newcommand{\rolecell}[1]{{\raggedright #1\par}}
\begin{tabular}{@{}p{0.19\linewidth}p{0.38\linewidth}p{0.38\linewidth}@{}}
\toprule
\rolecell{LLM role} & \rolecell{Inputs} & \rolecell{Outputs} \\[3pt]
\midrule
\rolecell{Perception planning} & \rolecell{Bounded series view, question context, eligible tools, Memory summaries, and execution limits.} & \rolecell{Tool names and arguments; a Memory query and retrieval choices.} \\[3pt]
\rolecell{TPS synthesis} & \rolecell{Tool evidence, retrieved Memory entries, series identity, channels, length, and question context.} & \rolecell{Measured fields to include and a factual summary $h_j$ describing the series.} \\[3pt]
\rolecell{Initial hypothesis} & \rolecell{Raw numerical series, question context, answer choices, and output requirements.} & \rolecell{A parsed candidate answer and its generated response, stored as an initial hypothesis in a named extension.} \\[3pt]
\rolecell{Initial extension planning} & \rolecell{Core findings, question, evidence requirements, and available tools and Memory.} & \rolecell{Analysis requests specifying operations, arguments, and required evidence.} \\[3pt]
\rolecell{Answerer} & \rolecell{Question and answer requirements, TPS with the provisional hypothesis, evidence coverage, available operations, and request budget.} & \rolecell{A task-formatted prediction with evidence references, or a request specifying the additional evidence needed.} \\[3pt]
\bottomrule
\end{tabular}
\endgroup
\end{table}

\subsection{Evidence updates in the harness loop}

The implementation maintains the core fields and accumulates named extensions through the update in Eq.~(\ref{eq:tps-update}).
\begin{itemize}
\item \textbf{Specify the need.} A request identifies the series, operations, arguments, and evidence types needed for further reasoning.
\item \textbf{Execute and attach.} The harness validates the request against available operations and the request budget, executes the analysis, and attaches the results with their sources to a new extension.
\item \textbf{Reassess the state.} Coverage is computed from available fields and returned evidence types. The Answerer uses the updated TPS to continue reasoning or produce the task-formatted output. The extension's coverage status records whether the requested evidence types were obtained.
\end{itemize}

\paragraph{Runtime limits.}
\label{app:runtime-limits}
The source configuration gives the Perception Controller a view of at most 48 observations per series, while numerical tools use the original sequence. Base perception admits up to nine tools per series; the initial planner adds at most one task extension per question. The Answerer can make one accepted evidence request within a three-step decision budget. Requested raw-series views are bounded to 128 points across the supplied channels. These limits are shared by the two conditions in the matched Tool Selector comparison.

\subsection{Successful executions on target benchmarks}
\label{app:harness-cases}

The following examples are drawn from completed Qwen3.8-27B trajectories using the TSRBench-developed harness. Both final predictions match the dataset references.

\paragraph{ChatTime: identifying a sustained level shift.}
The question asks whether a 128-point series has no obvious outlier, a sudden spike, or a level shift. The relevant evidence is compact:
\begin{itemize}
\item \textbf{Value profile:} the largest adjacent drop is $-2.48$, from index 31 to 32 (zero-based indexing).
\item \textbf{Temporal features:} the change-point tool detects a downward level shift at index 32; the trend tool reports a strong downward trend.
\item \textbf{Task extension:} an outlier-type analysis collects anomaly and change-point evidence for interpreting the observed change.
\end{itemize}
The Answerer combines the abrupt drop with the persistent change in level and selects \emph{C: level shift}. This execution illustrates the separation between locating and measuring a change in Perception and interpreting its type under the question's definitions.

\paragraph{TSAQA: reconstructing a temporal sequence.}
An electricity-demand question provides the first patch and four remaining patches, each containing 45 observations, and asks for their chronological order. A TPS extension evaluates how adjacent patches join. For patches $u$ and $v$, its boundary cost is
\begin{equation}
 c(u,v)=\frac{|u_{\mathrm{last}}-v_{\mathrm{first}}|
 +0.25\,|\Delta u_{\mathrm{end}}-\Delta v_{\mathrm{start}}|}{s+\epsilon},
\label{eq:appendix-boundary-cost}
\end{equation}
where $s$ is the standard deviation across all supplied patches, the two differences are the last and first adjacent increments, and $\epsilon=10^{-12}$. Candidate orders are ranked by the sum of boundary costs, including the join from the given first patch. The extension records a cost of 0.642 for \emph{B,C,A,D}, compared with 1.916 for the next candidate. Using this evidence, the Answerer revises its preliminary hypothesis \emph{D,A,B,C} to the correct sequence \emph{B,C,A,D}. The same TPS interface therefore supports both a categorical judgment in ChatTime and an ordered output in TSAQA.

\section{Tool Selector Training Details}
\label{app:selector}

This section specifies the feature representation, optimization, and deployment of the Tool Selector in Sec.~\ref{sec:selector}. The configuration below is used in the matched Qwen3.8-27B base-tool comparison. The selector refines each series' initial tool plan; the resulting evidence is processed by the same TPS construction and answering procedures described in Appendix~\ref{app:harness-implementation}.

\subsection{Policy inputs and network}

For candidate tool $i$ at decision $t$, the implementation constructs
\begin{equation}
\phi_{t,i}=[h(c_t);h(d_i);h(c_t)\odot h(d_i);v_{t,i}]
\in\mathbb R^{396},
\label{eq:appendix-tool-features}
\end{equation}
where $c_t$ combines the question with its task and dimension descriptors, $d_i$ contains the tool name, description, supported operations, and output fields, and $\odot$ is elementwise multiplication. The function $h$ maps lowercase text tokens into a fixed 128-dimensional signed hash vector and normalizes its length. The three text blocks therefore contribute 384 dimensions. The remaining 12 scalar features comprise:
\begin{itemize}
\item \textbf{Plan and tool information:} membership in the Perceiver proposal, proposal and candidate-set sizes, declared tool cost and reliability, multivariate applicability, text-vector similarity, and a constant feature.
\item \textbf{Current-series information:} scaled sequence length and channel count, standardized endpoint change, and first-difference variability relative to the series standard deviation. The latter two are bounded with $\tanh$.
\end{itemize}
These features provide a compact observation of the decision state in Eq.~(\ref{eq:tool-mdp}). A shared multilayer perceptron, $396\rightarrow64\rightarrow1$ with a $\tanh$ hidden activation, outputs the adjustment $f_\theta(\phi_{t,i})$ for each candidate.

\subsection{Perceiver prior and subset sampling}

The initial inclusion probability is 0.8 for a tool proposed by the Perceiver and 0.2 for another eligible tool. Zero initialization of the final linear layer makes the initial policy follow these priors. The learned adjustment is added to the prior logit, and the resulting logit is bounded to $[-8,8]$ before applying the sigmoid.

For the inclusion probabilities $p_{t,i}$ defined in Sec.~\ref{sec:selector}, conditioning on a nonempty subset gives
\begin{equation}
\pi_\theta(a_t\mid s_t)=
\frac{\prod_{i\in\mathcal T_t}p_{t,i}^{\mathbf 1[i\in a_t]}(1-p_{t,i})^{\mathbf 1[i\notin a_t]}}
{1-\prod_{i\in\mathcal T_t}(1-p_{t,i})},\qquad a_t\ne\varnothing.
\label{eq:tool-policy}
\end{equation}
The denominator removes the empty subset and normalizes the remaining probabilities.

Training samples tool memberships from this joint policy, resampling an empty subset. The eligible pool fits within the nine-tool base-perception budget, so sampled subsets retain their full membership. In efficiency-oriented deployment, tools with inclusion probability at least 0.5 are executed; if none passes, the highest-probability tool is selected. Retained tools preserve their Perceiver-provided arguments and order, while additional tools use their registered defaults.

\subsection{Deployment configurations}
\label{app:selector-deployment}

The deployment configurations use the deterministic thresholding policy from Sec.~\ref{sec:selector} with the same source-selected checkpoint. The efficiency-oriented configuration applies the inclusion rule directly. Accuracy-oriented deployment moderates the learned adjustment and applies evidence-preservation constraints after thresholding. Its inclusion probabilities retain the Perceiver prior:
\begin{equation}
p_{t,i}^{(\alpha)}=\sigma\!\left(\operatorname{clip}\!\left[
\operatorname{logit}(p_{t,i}^{0})+\alpha f_\theta(\phi_{t,i}),-8,8\right]\right),
\qquad \alpha=0.4.
\label{eq:selector-calibrated-inference}
\end{equation}
Here $p_{t,i}^{0}$ is the prior inclusion probability. Scaling the learned adjustment preserves the policy's tool-dependent preferences while reducing departures from the initial plan; $\alpha$ is not a fraction of tools to execute. No policy weights are updated during this adjustment.

After thresholding, the accuracy-oriented rule retains at least two eligible Perceiver-proposed tools, or all of them if fewer than two were proposed. A proposed \texttt{value\_profile} is retained; when both \texttt{volatility} and \texttt{statistics} are proposed, both are retained. These constraints preserve basic numerical evidence while leaving the learned selector to filter the remaining operations. The deployment rules are fixed before target evaluation and apply uniformly across target questions, without target-side tuning or reference-answer feedback.

In the accuracy-oriented matched comparison, both conditions explicitly present task-relevant TPS measurements and instruct the Answerer to check its initial judgment against them. Evidence matching recognizes equivalent field names, and interval slicing preserves both observations of a two-point interval. Holding these procedures fixed evaluates the learned selector together with its conservative deployment constraints against the no-selector Controller. Across deployment configurations, inference rules and evidence processing can differ; their comparison characterizes deployment choices rather than isolating the effect of a single selector parameter.

\subsection{Rollout reward and optimization}

Each training question generates $G=4$ complete harness rollouts. A rollout can contain several selection decisions when the question includes multiple series. For rollout $g$, the terminal reward is the correctness--cost objective in Eq.~(\ref{eq:tool-reward}), with $\lambda_{\mathrm{tool}}=0.01$ and $N_{\mathrm{base}}$ summed across its series. Its advantage is
\begin{equation}
\widehat A_g=R_g-\frac{1}{G-1}\sum_{g'\ne g}R_{g'},
\label{eq:appendix-tool-advantage}
\end{equation}
using the other rollouts of the same question as the baseline. Advantages retain the original reward scale, preserving the relative weight of correctness and tool cost.

Let $T_g$ be the number of selection decisions in rollout $g$. The likelihood ratio includes every selected subset in that rollout:
\begin{equation}
\rho_g(\theta)=\prod_{t=1}^{T_g}
\frac{\pi_\theta(a_{g,t}\mid s_{g,t})}
{\pi_{\mathrm{old}}(a_{g,t}\mid s_{g,t})}.
\label{eq:appendix-tool-ratio}
\end{equation}
Each subset probability includes the nonempty-set normalization in Eq.~(\ref{eq:tool-policy}). The implementation minimizes
\begin{equation}
\mathcal L(\theta)=\mathbb E_g\!\left[
-\min\!\left(\rho_g\widehat A_g,
\operatorname{clip}(\rho_g,1-\epsilon,1+\epsilon)\widehat A_g\right)
+\beta\overline D_g-\eta\overline H_g\right],
\label{eq:appendix-tool-loss}
\end{equation}
where $\overline D_g$ averages the KL divergence between each candidate's Bernoulli inclusion probability and its Perceiver prior, and $\overline H_g$ averages the corresponding Bernoulli entropies. These regularizers operate on the individual inclusion distributions; the policy-gradient term uses the conditioned subset probabilities. Table~\ref{tab:selector-training} gives the training settings.

\begin{table}[ht]
\centering\small
\caption{Configuration of the learned Tool Selector in the matched base-tool experiment. The LLM and source-developed Tool and Memory resources remain fixed during policy optimization.}
\label{tab:selector-training}
\begin{tabular}{@{}ll@{}}
\toprule
Setting & Value \\
\midrule
Source training / development questions & 96 / 48 \\
Training passes & 1 \\
Rollouts per question / questions per batch & 4 / 4 \\
Text-hash dimension / scalar features & $128$ per block / $12$ \\
Policy network & $396\rightarrow64\rightarrow1$, $\tanh$ \\
Perceiver prior: proposed / other tools & $0.8$ / $0.2$ \\
Optimizer / learning rate & Adam / $3\times10^{-3}$ \\
Update epochs per rollout batch & 4 \\
Policy-ratio clip $\epsilon$ & $0.2$ \\
Prior KL weight $\beta$ & $0.02$ \\
Entropy weight $\eta$ & $0.001$ \\
Gradient norm cap & $1.0$ \\
Cost weight $\lambda_{\mathrm{tool}}$ & $0.01$ per base-tool execution \\
Deployment inclusion threshold & $0.5$ \\
Random seed & 20260909 \\
\bottomrule
\end{tabular}
\end{table}

\subsection{Checkpoint selection and matched evaluation}

We evaluate the predetermined halfway and final checkpoints on the source development questions. Checkpoints are ranked first by answer accuracy, then by mean correctness--cost reward, with an earlier checkpoint preferred if both tie. The final checkpoint is selected in the reported experiment. Deployment uses its deterministic inclusion rule.

The Controller condition executes the \textsc{Perceiver}'s validated proposal without a selector; the RL condition applies the learned policy to refine that proposal. Both conditions use the same eligible tools, frozen backbone, Memory, extension rules, and evidence-request budget. Identical model requests share cached responses within the comparison, fixing model outputs for repeated prompts while allowing changed evidence to produce different requests. In this efficiency-oriented TSRBench comparison, Controller and selector accuracies are 61.89\% and 59.95\%, respectively. The 62.62\% harness result in Table~\ref{tab:transfer} comes from the separate main evaluation. Figure~\ref{fig:tool-efficiency}(b) retains the execution distribution from this matched comparison.

The efficiency-oriented cross-dataset evaluation freezes the selected checkpoint and TSRBench-developed Tool and Memory resources and deploys them on 412 questions from each target benchmark. Target inference uses the same Qwen3.8-27B backbone with temperature 0, an 8,192-token context limit, and at most 768 generated tokens per model call. Target accuracy is evaluated with response caching disabled, using the completed-run scoring rule in Appendix~\ref{app:metrics}. Figure~\ref{fig:tool-efficiency}(a) aggregates all five target benchmarks; the TSRBench comparison above provides the separate per-question distribution in panel (b).

The accuracy-oriented comparison covers the same five target benchmarks: Time-MQA, ChatTime, CaTSBench, TSAQA, and TimeOmni. Its no-selector control uses the same evidence-processing procedures specified in Appendix~\ref{app:selector-deployment}. Appendix~\ref{app:selector-operating-points} reports the endpoint and intermediate results, with costs normalized to the corresponding matched Controller conditions.

\paragraph{Execution accounting.}
Following Eq.~(\ref{eq:tool-reward}), the reward penalizes base-tool executions to reduce numerical analysis overhead while retaining answer correctness. Evaluation records all executed tools, summed over every series in a question:
\begin{equation}
N_{\mathrm{all}}=N_{\mathrm{base}}+N_{\mathrm{question}}
+N_{\mathrm{planned}}+N_{\mathrm{requested}}.
\label{eq:appendix-tool-counts}
\end{equation}
The additional terms count question-specific analytical tools, initial TPS-extension tools, and tools triggered by accepted Answerer requests. An evidence revisit counts an accepted request, which can execute multiple operations. Initial-hypothesis generation is counted as a model call. Logical model-call counts include cached responses and are tracked separately from tool executions; a tool that invokes the LLM contributes to both categories. Appendix~\ref{app:tool-analysis} provides the corresponding empirical breakdowns.

\section{Supplementary Experimental Results}
\label{app:supp-results}
\label{app:tool-analysis}

We supplement Sec.~\ref{sec:experiments} with results across task categories, answer revisions after evidence integration, and the distribution of tool executions. Task-level scores use the same completed predictions as Table~\ref{tab:transfer}; the Tool Selector analysis uses the separate matched comparison specified in Appendix~\ref{app:selector}.

\subsection{Performance across capabilities and task types}
\label{app:task-results}

Tables~\ref{tab:supp-capabilities} and~\ref{tab:supp-target-tasks} compare the complete harness with direct zero-shot answering under each backbone. Benchmark-native categories preserve differences in analytical demands. Time-MQA is grouped by answer format because its detailed task labels are heterogeneous. Its evaluation subset follows the sampling procedure in Appendix~\ref{app:protocol}.

\paragraph{Source capabilities.}
On TSRBench, the Qwen-based harness improves all four capability groups. The largest increases occur in reasoning (36.26\% to 65.50\%) and decision-making (25.40\% to 50.79\%). The GPT-based harness also improves reasoning, perception, and decision-making, while prediction decreases from 61.11\% to 57.41\%. These results locate the aggregate source gain across distinct analytical demands.

\begin{table}[htbp]
\centering\small
\caption{Accuracy (\%) across source capabilities and target question categories, using the matched same-backbone predictions underlying Table~\ref{tab:transfer}. TSRBench is grouped by capability, ChatTime and TimeOmni-1 by task, and Time-MQA by answer format. $\Delta$ is \modelname{} minus zero-shot in percentage points; bold indicates the higher score within each pair.}
\label{tab:supp-capabilities}
\begingroup
\setlength{\tabcolsep}{4pt}
\renewcommand{\arraystretch}{1.07}
\begin{tabular}{@{}lrrrrrr@{}}
\toprule
& \multicolumn{3}{c}{Qwen3.8-27B} & \multicolumn{3}{c}{GPT-5.6-sol} \\
\cmidrule(lr){2-4}\cmidrule(l){5-7}
Capability / task & Zero-shot & \modelname{} & $\Delta$ & Zero-shot & \modelname{} & $\Delta$ \\
\midrule
\multicolumn{7}{l}{\textbf{TSRBench}} \\
Decision-making & 25.40 & \textbf{50.79} & +25.40 & 44.44 & \textbf{53.97} & +9.52 \\
Perception & 68.57 & \textbf{71.43} & +2.86 & 82.86 & \textbf{87.14} & +4.29 \\
Prediction & 50.00 & \textbf{59.26} & +9.26 & \textbf{61.11} & 57.41 & -3.70 \\
Reasoning & 36.26 & \textbf{65.50} & +29.24 & 71.93 & \textbf{81.87} & +9.94 \\
\midrule
\multicolumn{7}{l}{\textbf{ChatTime}} \\
Outliers & 92.10 & \textbf{96.20} & +4.10 & 96.20 & \textbf{98.90} & +2.70 \\
Volatility & 60.40 & \textbf{97.00} & +36.60 & \textbf{97.60} & 97.50 & -0.10 \\
Trend & 83.30 & \textbf{100.00} & +16.70 & 98.40 & \textbf{100.00} & +1.60 \\
Seasonality & 60.70 & \textbf{91.20} & +30.50 & 66.70 & \textbf{91.10} & +24.40 \\
\midrule
\multicolumn{7}{l}{\textbf{Time-MQA}} \\
Multiple choice & 62.10 & \textbf{64.92} & +2.82 & 59.68 & \textbf{64.11} & +4.44 \\
True/false & 51.83 & \textbf{67.07} & +15.24 & 64.63 & \textbf{72.56} & +7.93 \\
\midrule
\multicolumn{7}{l}{\textbf{TimeOmni-1}} \\
Causality discovery & 40.34 & 40.34 & 0.00 & 51.70 & \textbf{52.84} & +1.14 \\
Scenario understanding & 74.72 & \textbf{83.71} & +8.99 & \textbf{97.19} & 94.94 & -2.25 \\
Decision-making & 10.34 & \textbf{12.07} & +1.72 & \textbf{82.76} & 79.31 & -3.45 \\
\bottomrule
\end{tabular}
\endgroup
\end{table}

\paragraph{Transfer across question types.}
The target results reveal gains in both temporal perception and task-specific interpretation:
\begin{itemize}
\item \textbf{Temporal properties.} On ChatTime, Qwen improves seasonality recognition from 60.70\% to 91.20\% and volatility recognition from 60.40\% to 97.00\%. GPT improves seasonality from 66.70\% to 91.10\%; both harnesses reach 100\% on trend questions.
\item \textbf{Relational and structured judgments.} On TSAQA, Qwen improves temporal relationships from 13.24\% to 45.59\% and data transformation from 39.71\% to 55.88\%. On CaTSBench, its peak-timing comparison rises from 70.59\% to 91.18\%, and volatility comparison from 76.47\% to 88.24\%.
\item \textbf{Contextual interpretation.} On TimeOmni-1, Qwen improves scenario understanding from 74.72\% to 83.71\%, accounting for most of its aggregate gain. GPT starts at 97.19\% on this category and reaches 94.94\% with the harness, illustrating the dependence of transfer gains on the backbone's initial capability.
\end{itemize}
The breakdowns connect the transfer results to the statistical, temporal, and task-specific information represented in TPS. Improvements span several question types, while categories such as TSAQA comparison and perturbed-series retrieval show that the benefit is not uniform.

\begin{table}[htbp]
\centering\small
\caption{Task-level accuracy (\%) on CaTSBench and TSAQA. All evaluated task categories are included. Both conditions use the same questions within each benchmark. Bold and $\Delta$ follow Table~\ref{tab:supp-capabilities}.}
\label{tab:supp-target-tasks}
\begingroup
\setlength{\tabcolsep}{4pt}
\renewcommand{\arraystretch}{1.07}
\begin{tabular}{@{}lrrrrrr@{}}
\toprule
& \multicolumn{3}{c}{Qwen3.8-27B} & \multicolumn{3}{c}{GPT-5.6-sol} \\
\cmidrule(lr){2-4}\cmidrule(l){5-7}
Capability / task & Zero-shot & \modelname{} & $\Delta$ & Zero-shot & \modelname{} & $\Delta$ \\
\midrule
\multicolumn{7}{l}{\textbf{CaTSBench}} \\
Caption retrieval: cross-domain & 100.00 & 100.00 & 0.00 & 100.00 & 100.00 & 0.00 \\
Caption retrieval: perturbed & 100.00 & 100.00 & 0.00 & 100.00 & 100.00 & 0.00 \\
Caption retrieval: same-domain & 100.00 & 100.00 & 0.00 & 100.00 & 100.00 & 0.00 \\
Paraphrase consistency & 100.00 & 100.00 & 0.00 & 100.00 & 100.00 & 0.00 \\
Series comparison: amplitude & 94.12 & \textbf{100.00} & +5.88 & 100.00 & 100.00 & 0.00 \\
Series comparison: mean & 97.06 & \textbf{100.00} & +2.94 & 100.00 & 100.00 & 0.00 \\
Series comparison: peak timing & 70.59 & \textbf{91.18} & +20.59 & 94.12 & \textbf{100.00} & +5.88 \\
Series comparison: phenomenon & 94.12 & \textbf{100.00} & +5.88 & 97.06 & 97.06 & 0.00 \\
Series comparison: volatility & 76.47 & \textbf{88.24} & +11.76 & 97.06 & \textbf{100.00} & +2.94 \\
Series retrieval: cross-domain & 100.00 & 100.00 & 0.00 & 100.00 & 100.00 & 0.00 \\
Series retrieval: perturbed & \textbf{82.35} & 79.41 & -2.94 & \textbf{97.06} & 94.12 & -2.94 \\
Series retrieval: same-domain & 100.00 & 100.00 & 0.00 & 100.00 & 100.00 & 0.00 \\
\midrule
\multicolumn{7}{l}{\textbf{TSAQA}} \\
Anomaly detection & 49.28 & 49.28 & 0.00 & 44.93 & 44.93 & 0.00 \\
Characterization & \textbf{84.06} & 81.16 & -2.90 & 59.42 & \textbf{62.32} & +2.90 \\
Classification & 44.93 & \textbf{50.72} & +5.80 & 50.72 & 50.72 & 0.00 \\
Comparison & \textbf{81.16} & 73.91 & -7.25 & \textbf{62.32} & 59.42 & -2.90 \\
Data transformation & 39.71 & \textbf{55.88} & +16.18 & 69.12 & \textbf{73.53} & +4.41 \\
Temporal relationships & 13.24 & \textbf{45.59} & +32.35 & 60.29 & 60.29 & 0.00 \\
\bottomrule
\end{tabular}
\endgroup
\end{table}

\subsection{Answer revisions after evidence integration}
\label{app:answer-revisions}

This section supplements the pipeline analysis in Sec.~\ref{sec:pipeline-analysis} with initial and final accuracies and evidence-request frequencies. The provisional hypothesis described in Sec.~\ref{sec:method} provides a reference for examining how judgments change after TPS evidence is incorporated. For each completed trajectory, we score this hypothesis and the final prediction with the same benchmark parser. Let $\mathrm{Acc}_{\mathrm{initial}}$ and $\mathrm{Acc}_{\mathrm{final}}$ denote their accuracies in percent. The net change in percentage points is
\begin{equation}
\Delta_{\mathrm{revision}}
=\mathrm{Acc}_{\mathrm{final}}-\mathrm{Acc}_{\mathrm{initial}}.
\label{eq:revision-gain}
\end{equation}
Table~\ref{tab:supp-revisions} reports initial and final accuracies and their net change for each benchmark and backbone. Figure~\ref{fig:pipeline-revisions}(a) visualizes answer transitions and the net accuracy change for the GPT backbone. The 95\% intervals for the net change are the 2.5th and 97.5th percentiles of 20,000 question-level bootstrap resamples within each benchmark, preserving the initial--final prediction pairs. These intervals describe question-level variation within the evaluated subsets.

\begin{table}[htbp]
\centering\small
\caption{Accuracy before and after TPS evidence integration within the same harness trajectory. Initial Accuracy scores the recorded pre-TPS hypothesis; Final Accuracy scores the final prediction. Both are reported in percent. $\Delta$ is the net accuracy change in percentage points (pp), computed before rounding. Revisit gives the percentage of questions with an accepted evidence request. The initial hypothesis is a within-trajectory reference, distinct from the independently evaluated zero-shot baseline. Time-MQA uses the same subset as Table~\ref{tab:supp-capabilities}.}
\label{tab:supp-revisions}
\begin{tabular}{@{}lrrrr@{}}
\toprule
Benchmark & Initial Accuracy (\%) & Final Accuracy (\%) & $\Delta$ (pp) & Revisit (\%) \\
\midrule
\multicolumn{5}{l}{\textbf{Qwen3.8-27B}} \\
TSRBench & 44.17 & 62.62 & +18.45 & 0.00 \\
Time-MQA & 59.47 & 65.78 & +6.31 & 0.24 \\
ChatTime & 74.30 & 96.10 & +21.80 & 0.00 \\
CaTSBench & 93.45 & 96.60 & +3.16 & 0.00 \\
TSAQA & 52.18 & 59.47 & +7.28 & 0.24 \\
TimeOmni-1 & 51.21 & 55.10 & +3.88 & 0.00 \\
\midrule
\multicolumn{5}{l}{\textbf{GPT-5.6-sol}} \\
TSRBench & 62.86 & 72.09 & +9.22 & 19.90 \\
Time-MQA & 63.83 & 67.48 & +3.64 & 2.18 \\
ChatTime & 89.78 & 96.88 & +7.10 & 0.00 \\
CaTSBench & 99.03 & 99.27 & +0.24 & 0.97 \\
TSAQA & 61.17 & 58.50 & -2.67 & 11.41 \\
TimeOmni-1 & 75.00 & 74.76 & -0.24 & 48.54 \\
\bottomrule
\end{tabular}
\end{table}

The revisit column additionally records use of the evidence-request loop. GPT requests additional evidence on 19.90\% of TSRBench questions and 48.54\% of TimeOmni-1 questions, whereas Qwen mainly completes its judgments from the initial TPS. The same interface thus supports both direct reasoning over the initial state and further perception when requested.

With Qwen, final accuracy exceeds initial accuracy on every evaluated benchmark. On TSRBench, accuracy increases from 44.17\% to 62.62\%; on ChatTime, it increases from 74.30\% to 96.10\%; and on TSAQA, it increases from 52.18\% to 59.47\%. GPT also obtains net gains on TSRBench, Time-MQA, ChatTime, and CaTSBench, but its revisions on TSAQA and TimeOmni-1 yield small net losses. These comparisons characterize the complete evidence-integration process; they do not isolate the contribution of the TPS representation or additional loop iterations. Appendix~\ref{app:harness-cases} gives concrete successful executions.

\subsection{Efficiency-oriented execution across stages and capabilities}
\label{app:execution-results}

The original efficiency-oriented selector comparison uses Qwen3.8-27B on paired TSRBench questions with the resources, downstream procedures, and response-cache protocol specified in Appendix~\ref{app:selector}. Controller executes the \textsc{Perceiver}'s validated proposal without selection; RL applies the efficiency-oriented Tool Selector. These execution breakdowns do not describe the conservative \modelname{}$^{\star}$ configuration. All counts include executions across every input series, following Eq.~(\ref{eq:appendix-tool-counts}).

\begin{figure}[htbp]
\centering
\includegraphics[width=\textwidth]{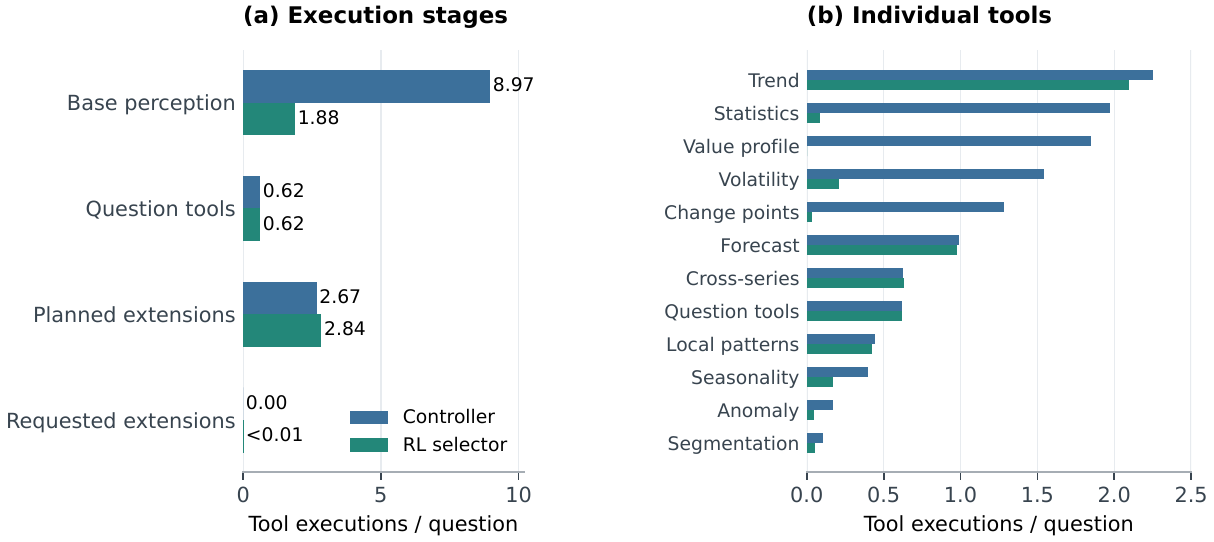}
\caption{Tool execution breakdown for the matched Controller and RL conditions. (a) Mean executions at each stage. (b) Mean executions by tool across all stages, ordered by Controller usage; the seven question-specific analytical and verification tools are pooled as Question tools. Both panels use executions per question; requested extensions average zero for Controller and less than 0.01 for RL.}
\label{fig:supp-tool-stages}
\end{figure}

\paragraph{Where execution savings occur.}
Figure~\ref{fig:supp-tool-stages} decomposes the reduction from 12.25 to 5.34 executions per question. Base perception accounts for the savings: its mean falls from 8.97 to 1.88. Question-specific tools remain at 0.62, while planned extensions increase from 2.67 to 2.84. Later evidence acquisition therefore remains active after a smaller initial tool set is executed. At the operator level, the largest reductions occur in statistics, value profiles, volatility, and change-point analysis. Trend and cross-series operations retain substantial execution volume, while question-specific verification tools retain their calls.

\begin{figure}[htbp]
\centering
\includegraphics[width=\textwidth]{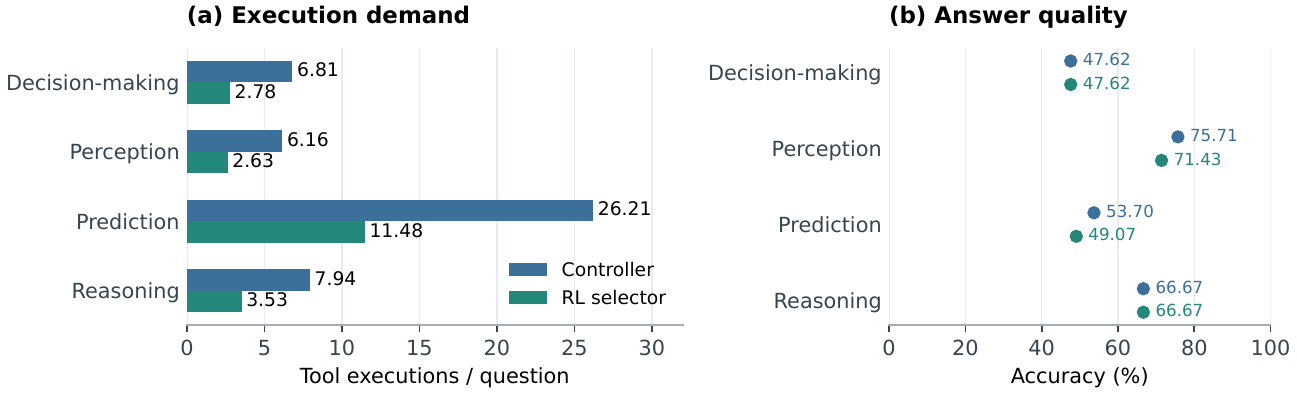}
\caption{Execution--accuracy trade-off across TSRBench capabilities. (a) Mean tool executions per question across all stages. (b) Accuracy on the same questions. The selector reduces execution demand in every group while preserving accuracy in decision-making and reasoning. Colors identify the same policies in both panels.}
\label{fig:supp-tool-capabilities}
\end{figure}

\paragraph{How the trade-off varies across capabilities.}
Figure~\ref{fig:supp-tool-capabilities} reports execution demand together with accuracy. Executions decrease in all four capability groups. Decision-making and reasoning retain their accuracies at 47.62\% and 66.67\%, respectively, with more than 55\% fewer executions. Perception and prediction exhibit accuracy decreases of 4.29 and 4.63 percentage points. Within prediction, numerical forecasting improves from 45.83\% to 50.00\% while event forecasting decreases from 69.44\% to 47.22\%. The cost--quality balance therefore varies with the evidence demands of the task, even under the same selection policy.

\paragraph{Distribution and measurement scope.}
The reduction is broadly distributed: executions decrease on 411 questions and remain unchanged on one. The median falls from six to three, the interquartile range from 5--9 to 2--4, and the 95th percentile from 38 to 17. The 56.39\% reduction in mean executions has a paired-bootstrap 95\% interval of 55.41--57.39\%, using 20,000 question-level resamples. Accuracy changes from 61.89\% to 59.95\%, a decrease of 1.94 percentage points. These statistics quantify an execution--accuracy trade-off under the fixed comparison. Accepted evidence revisits occur on zero Controller questions and one RL question; logical model-call totals are 3,393 and 3,392. These counts quantify the analytical work saved through tool selection; end-to-end runtime additionally depends on operation runtimes and LLM inference.

\subsection{Selector operating points and transfer scope}
\label{app:selector-operating-points}

Table~\ref{tab:selector-transfer-regimes} reports all five target benchmarks for the accuracy-oriented and efficiency-oriented configurations. Their mean accuracies are 74.71\% and 72.54\%, respectively, computed from the five benchmark scores. TSRBench is a source benchmark and is excluded from this target mean. The efficiency-oriented configuration additionally reaches 59.95\% on TSRBench; this source-domain score does not enter Figure~\ref{fig:tool-efficiency}(a).

\begin{table}[htbp]
\centering\small
\caption{Target accuracy (\%) for the two deployment regimes. Accuracy-oriented values correspond to \modelname{}$^{\star}$ in Table~\ref{tab:transfer}. Both regimes reuse the same learned checkpoint.}
\label{tab:selector-transfer-regimes}
\begin{tabular}{@{}lrr@{}}
\toprule
Benchmark & Accuracy-oriented & Efficiency-oriented \\
\midrule
Time-MQA & 65.78 & 63.67 \\
ChatTime & 96.12 & 94.90 \\
CaTSBench & 96.84 & 93.69 \\
TSAQA & 59.71 & 57.04 \\
TimeOmni & 55.10 & 53.40 \\
\midrule
Five-target mean & 74.71 & 72.54 \\
\bottomrule
\end{tabular}
\end{table}

\begin{table}[b]
\centering\small
\caption{Measured accuracy--cost trade-offs across all five target benchmarks, plotted in Figure~\ref{fig:tool-efficiency}(a). Tool savings equal 100\% minus relative execution cost; each cost is normalized to its matched Controller.}
\label{tab:selector-operating-points}
\begin{tabular}{@{}lrrr@{}}
\toprule
Configuration & Accuracy (\%) & Relative cost (\%) & Tool savings (\%) \\
\midrule
Efficiency-oriented & 72.54 & 43.61 & 56.39 \\
Intermediate 1 & 73.62 & 57.43 & 42.57 \\
Intermediate 2 & 74.10 & 74.28 & 25.72 \\
Intermediate 3 & 74.55 & 85.23 & 14.77 \\
Accuracy-oriented & 74.71 & 98.24 & 1.76 \\
\bottomrule
\end{tabular}
\end{table}

\paragraph{Relative execution cost.}
For each matched comparison, the no-selector Controller condition defines 100\% cost. Relative tool cost is the ratio of total executions, including every analysis stage and input series:
\begin{equation}
C=100\,\frac{\sum_{q\in\mathcal Q}N_{\mathrm{all}}^{\mathrm{selector}}(q)}
{\sum_{q\in\mathcal Q}N_{\mathrm{all}}^{\mathrm{Controller}}(q)}.
\label{eq:selector-relative-cost}
\end{equation}
Here $\mathcal Q$ contains the evaluated questions from all five target benchmarks. Costs are pooled over these questions, whereas accuracy is the unweighted mean of the five benchmark scores. Accuracy-oriented deployment retains 98.24\% of the matched Controller's executions, including a 3.5\% reduction on Time-MQA; efficiency-oriented deployment retains 43.61\%. Table~\ref{tab:selector-operating-points} gives all five measured configurations. Consistent with the execution penalty in Eq.~(\ref{eq:tool-reward}), relative cost measures tool use; wall-clock latency and LLM inference are separate quantities.

The five configurations demonstrate that a fixed learned selector can support different accuracy--cost balances through its deployment rules. More conservative evidence acquisition sustains higher accuracy, while stronger filtering reduces tool execution. Inference rules and evidence processing differ across configurations, as discussed in Appendix~\ref{app:selector-deployment}; the comparison evaluates complete deployment settings rather than an isolated selector parameter.

\end{document}